%% file: arxiv.tex
\documentclass[letterpaper]{article} 
\usepackage{aaai24}  
\usepackage{times}  
\usepackage{helvet}  
\usepackage{courier}  
\usepackage[hyphens]{url}  
\usepackage{graphicx} 
\usepackage{natbib}  
\usepackage{caption} 
\usepackage{algorithm}
\usepackage{algorithmic}

\usepackage{newfloat}
\usepackage{listings}
\DeclareCaptionStyle{ruled}{labelfont=normalfont,labelsep=colon,strut=off} 
\floatstyle{ruled}
\newfloat{listing}{tb}{lst}{}
\floatname{listing}{Listing}
\usepackage{amsmath,amssymb,amsfonts}
\usepackage{textcomp}
\usepackage{subcaption}
\usepackage[inline]{enumitem}

\usepackage{mathtools}
\usepackage{threeparttable}
\usepackage{fancyhdr} 
\usepackage[nolist]{acronym} 
\usepackage{comment}
\usepackage{booktabs}
\usepackage{bm} 
\usepackage{multirow}
\usepackage[utf8]{inputenc}
\usepackage[title]{appendix} 
\usepackage{derivative}
\usepackage{clipboard} 

\include{acronyms}

\title{Exploiting Symmetric Temporally Sparse BPTT for Efficient RNN Training}

\author{
    Xi Chen\textsuperscript{\rm 1},
    Chang Gao\textsuperscript{\rm 2},
    Zuowen Wang\textsuperscript{\rm 1},
    Longbiao Cheng\textsuperscript{\rm 1},
    Sheng Zhou\textsuperscript{\rm 1},\\
    Shih-Chii Liu\textsuperscript{\rm 1},
    Tobi Delbruck\textsuperscript{\rm 1}
}
\affiliations{
    \textsuperscript{\rm 1}Sensors Group, Institute of Neuroinformatics, University of Zurich and ETH Zurich\\
    \textsuperscript{\rm 2}Department of Microelectronics, Delft University of Technology\\
    \textsuperscript{\rm 1}\texttt{\{xi, zuowen, longbiao, shengzhou, shih, tobi\}@ini.uzh.ch}\\
    \textsuperscript{\rm 2}\texttt{chang.gao@tudelft.nl}\\
}

\begin{document}

\maketitle

\begin{abstract}
Recurrent Neural Networks (RNNs) are useful in temporal sequence tasks.
However, training RNNs involves dense matrix multiplications which require hardware that can support a large number of arithmetic operations and memory accesses. Implementing online training of RNNs on the edge 
calls for optimized algorithms for an efficient deployment on hardware.
Inspired by the spiking neuron model, 
the Delta RNN 
exploits temporal sparsity during inference by skipping over the update of hidden states from those inactivated neurons whose change of activation across two timesteps is below a defined threshold.
This work describes a training algorithm for Delta RNNs that exploits temporal sparsity in the backward propagation phase to reduce computational requirements for training on the edge. Due to the symmetric computation graphs of forward and backward propagation during training, the gradient computation of inactivated neurons can be skipped.
Results show a reduction of \(\sim\)80\% in matrix operations for training a 56k parameter Delta LSTM on the Fluent Speech Commands dataset with negligible accuracy loss.
Logic simulations of a hardware accelerator designed for the training algorithm show 2-10X speedup in matrix computations for an activation sparsity range of 50\%-90\%.
Additionally, we show that the proposed Delta RNN training will be useful for online incremental learning on edge devices with limited computing resources. 
\end{abstract}

\section{Introduction}
\label{sec:intro}



\tips{rnn} are widely used in applications involving temporal sequence inputs such as edge audio voice wakeup, keyword spotting, and spoken language understanding.
These RNNs are commonly trained once and then deployed, but there is an opportunity to continually improve their accuracy and classification power without giving up privacy by incremental training on edge devices.
Training of RNNs on the edge requires a hardware platform that has enough computing resources and memory to support the large number of arithmetic operations and data transfers. This is because the computation in RNNs consists mainly of \tips{mxv}, which is a memory-bounded operation. 
An effective method to reduce the energy consumption for training RNNs is to minimize the number of memory accesses.

Among various approaches that exploit sparsity in RNN inference to improve efficiency~\cite{Kadetotad2020-seo-lstm-cg-sparsity-jssc,Srivastava2019-Seo-structured-RNN-quantization,Chen2022-cong-lstm-sparse-bit-shift-macs,Lindmar2022-liu-targeted-dropout-lstm-training}, a previously proposed biologically inspired network model named \tip{dn} ~\cite{neil2016delta}, uses temporal sparsity to dramatically reduce memory access and \tip{mac} operations during inference. 
By introducing a delta threshold on neuron activation changes, the update of slow-changing activations can be skipped, thus saving a large number of computes while achieving comparable accuracy.
Hardware inference accelerators that exploit this temporal sparsity~\cite{GaoDeltaRNN2018, edgedrnn, gao2022spartus} can achieve 5-10X better energy efficiency with a custom design architecture that performs zero-skipping on sparse delta vectors. However, these accelerators only do inference, i.e., the forward propagation. 
This paper proves for the first time that the identical forward delta sparsity can be used in the backward propagation of training RNNs without extra accuracy loss and extends the \tip{dn} framework to the entire training process. The main contributions of this work are:
\begin{enumerate}
    \item
    The first mathematical formulation for Delta RNN training, showing that Delta RNN training is inherently a type of sparse \tip{bptt}, utilizing the identical temporal sparsity during both forward and backward propagation. Moreover, due to this consistent temporal sparsity, any speed improvements seen during Delta RNN inference can also be observed during the training of these networks. 
    \item
    Empirical results showing that for a fixed number of training epochs, a Delta RNN training uses 7.3X fewer training operations compared to the dense RNN with only a factor of 1.16X increase in error rate on the \tip{fscd}.
    \item 
    Empirical results showing that on the frequently used \tip{gscd} used for edge keyword spotting, \(\sim\)80\% training operations can be saved in an incremental learning setting.
    \item \tip{rtl} simulation results of the first hardware accelerator designed for training Delta RNNs which can achieve 2-10X speedup for an activation sparsity range of 50\%-90\%.
\end{enumerate}

The rest of the paper is organized as follows. Section~\ref{sec:method} describes the \tip{dn} theory for RNN training and shows the theoretical reduction in computation and memory access. Section~\ref{sec:experiments} describes the methodology for performance evaluation of Delta RNNs and shows experimental results. Section~\ref{sec:related_works} compares this work with others.
Section~\ref{sec:conclusion} presents concluding remarks.

\section{Methodology}
\label{sec:method}

This section summarizes the key concepts of the Delta Network, extends the theory to the \tip{bptt} process of RNN training, and shows its theoretical reduction in computation costs.

\begin{figure}[t]
    \centering
    \begin{subfigure}[t]{1\columnwidth}
        \centering
        \includegraphics[width=0.6\textwidth]{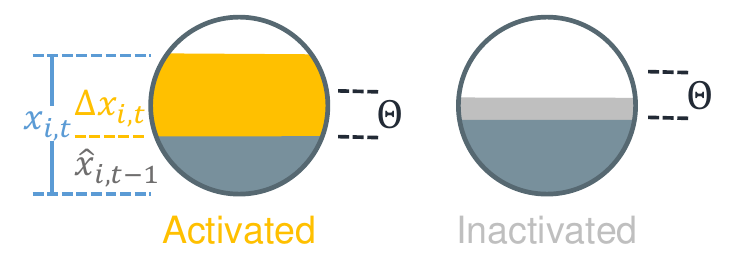}
        \caption{Delta neuron is activated when the activation change \(|\Delta|\) exceeds the delta threshold \(\Theta\) at a certain timestep.}
        \label{fig:dn_neuron}
    \end{subfigure}
    \begin{subfigure}[t]{1\columnwidth}
        \centering
        \includegraphics[width=0.6\textwidth]{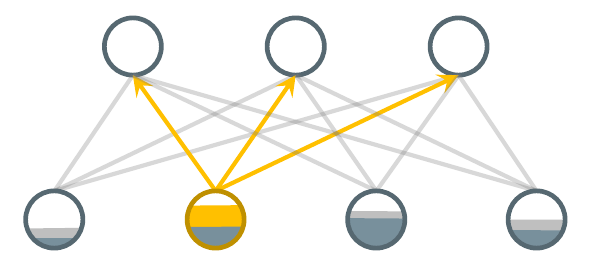}
        \caption{Sparse forward propagation: Activated neurons propagate their \(\Delta\) to the next layer, while other neurons remain unchanged.}
        \label{fig:dn_fp}
    \end{subfigure}
    \begin{subfigure}[t]{1\columnwidth}
        \centering
        \includegraphics[width=0.6\textwidth]{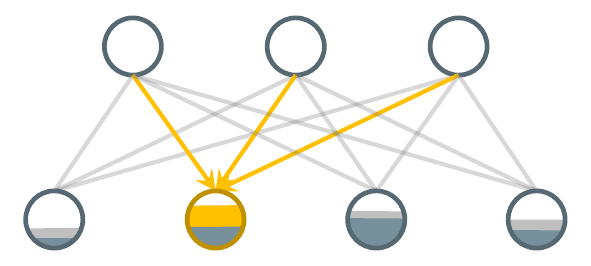}
        \caption{Sparse backward propagation: Only these neurons need to transmit their \(\Delta\) errors at that timestep in the backward phase.}
        \label{fig:dn_bp}
    \end{subfigure}
    \caption{Delta network concept for vanilla RNN with recurrent connections omitted.}
    \label{fig:dn_concept}
\end{figure}

\subsection{Delta Network Formulation}
\label{subsec:dn_form}

\begin{figure}[t]
    \centering
    \begin{subfigure}[t]{1\columnwidth}
        \centering
        \includegraphics[width=1\textwidth]{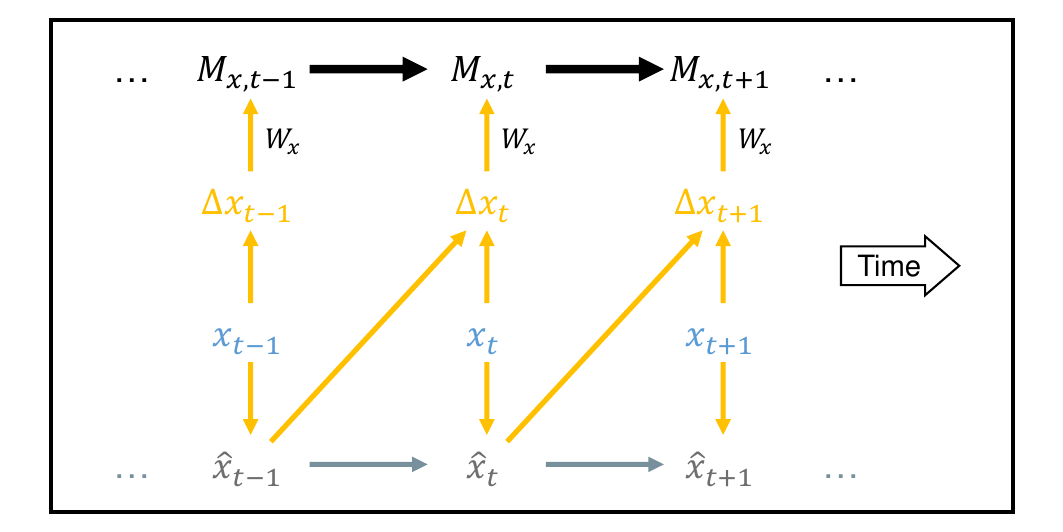}
        \caption{Forward computation graph illustrating input \(x_t\).}
        \label{fig:cg_fp}
    \end{subfigure}
    \begin{subfigure}[t]{1\columnwidth}
        \centering
        \includegraphics[width=1\textwidth]{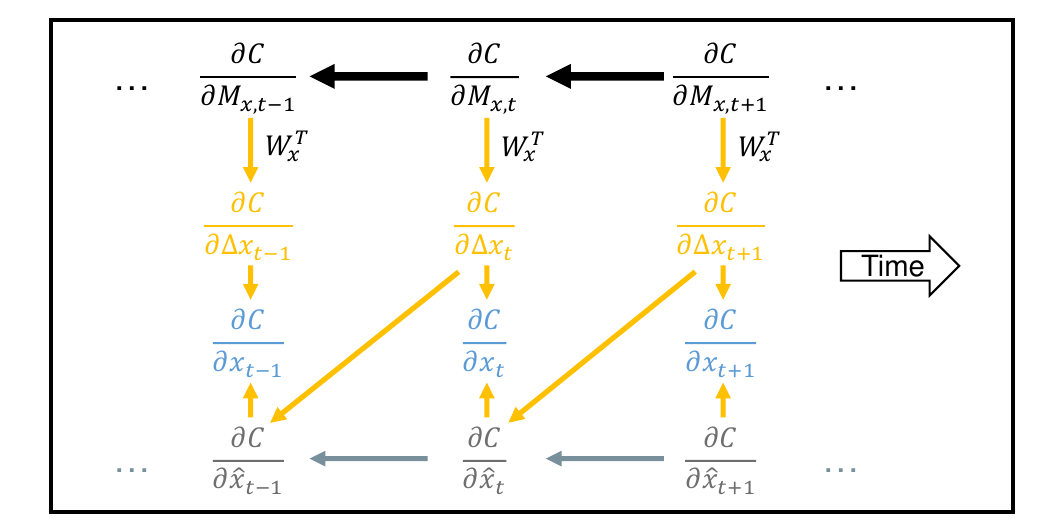}
        \caption{Backward computation graph illustrating input \(x_t\).}
        \label{fig:cg_bp}
    \end{subfigure}
    \caption{Inference and training computation graphs of vanilla Delta RNNs illustrating input \(x_t\). Graphs illustrating hidden states \(h_t\) are similar. Gold and grey arrows denote the paths for activated and inactivated neurons respectively, and black arrows include paths for all neurons.}
    \label{fig:cg_dn}
\end{figure}

\begin{figure}[t]
    \centering
    \includegraphics[width=0.9\columnwidth]{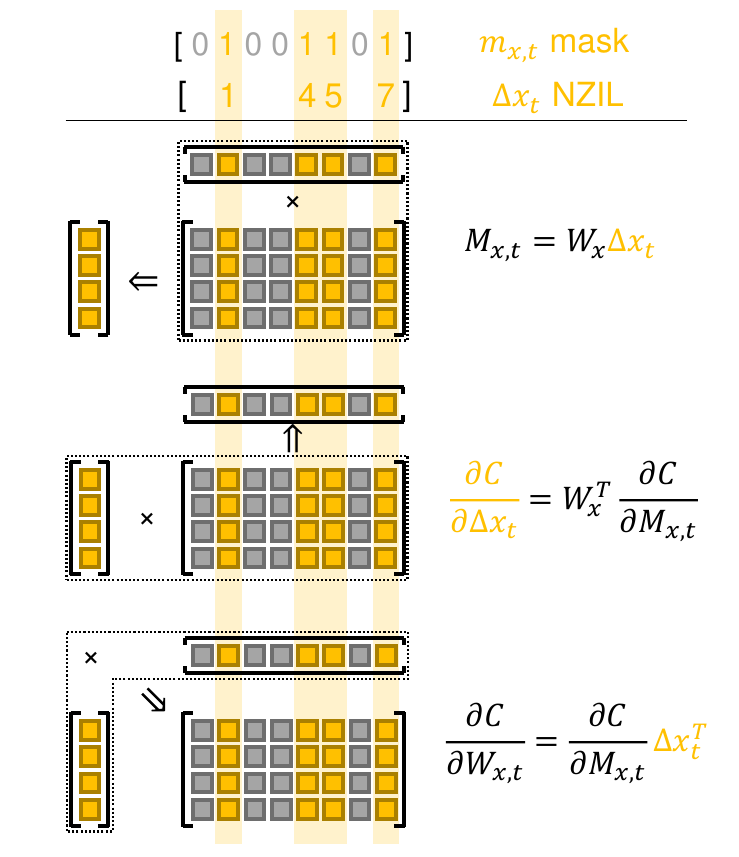}
    \caption{Sparse MxV in training Delta RNNs.}
    \label{fig:3_sparse_mxv}
\end{figure}

In a vanilla RNN layer, the pre-activation vector \(Z_{t}\) and the hidden state vector \(h_{t}\) at time step \(t\) are given by:
\begin{equation}
\label{eq:eq1}
Z_{t} = W_{x} x_{t} + W_{h} h_{t-1} + b_{h}
\end{equation}
\begin{equation}
\label{eq:eq2}
h_{t} = \tanh ( Z_{t} )
\end{equation}
where \(W_{x}\), \(W_{h}\) are the weight matrices for input and hidden states respectively,
\(x_t\) is the input vector, 
and \(b_{h}\) is the bias vector.
In the \tip{dn} formulation, these are calculated recursively by adding a new state variable vector, $M_t$ holding a preactivation memory:
\begin{equation} \label{eq:m_t}
M_{t} = W_{x} {\Delta x}_{t} + W_{h} {\Delta h}_{t-1} + M_{t-1}
\end{equation}
\begin{equation}
h_{t} = \tanh( M_{t} )
\end{equation}
if we define delta vectors \({\Delta x}\) and \({\Delta h}\) as:
\begin{equation}
{\Delta x}_{t} = x_{t} - x_{t-1}
\end{equation}
\begin{equation}
{\Delta h}_{t} = h_{t} - h_{t-1}
\end{equation}
with the initial state \(M_{0} = b_{h}\), where \(M_{t-1}\) stores the pre-activation from the previous time step. 

Fig.~\ref{fig:cg_fp} illustrates the $x_t$ part of these inference equations as a graph to clarify the steps and data dependencies.
It shows how $\hat{x}_{t-1}$ and $x_t$ combine to form $\Delta x_t$ and how it is multiplied by $W_x$ to compute $M_{x,t}$. 

Since the internal states of an RNN have temporal stability, 
the \tip{dn} zeros out small changes in activations and hidden states, i.e. , the \tip{dn} treats the values in \({|\Delta x_{t}}|\) and \({|\Delta h_{t}}|\) as zeros if they are smaller than a given delta threshold \(\Theta\), and those neurons are considered as “inactivated” (Fig.~\ref{fig:dn_neuron}). Then the \tip{dn} only propagates the changes of those activated neurons (Fig.~\ref{fig:dn_fp}), while the inactivated neurons keep current states until their changes go over the threshold later.

Formally, \(\hat{x}_{i,t}\) denotes the latest value of the \(i\)-th element of the input vector at the \(t\)-th time step. The values \(\hat{x}_{i,t}\) and \({\Delta x}_{i,t}\) will only be updated if the absolute difference between the current input \(x_{i,t}\) and the previously stored state \(\hat{x}_{i,t-1}\) is larger than the delta threshold \(\Theta\):
\begin{equation}
\hat{x}_{i,t} =
\begin{cases}
x_{i,t},         & \text{if } \lvert x_{i,t} - \hat{x}_{i,t-1} \rvert > \Theta \\
\hat{x}_{i,t-1}, & \text{otherwise}
\end{cases}
\end{equation}
\begin{equation} \label{eq:x_del_it}
{\Delta x}_{i,t} =
\begin{cases}
x_{i,t} - \hat{x}_{i,t-1}, & \text{if } \lvert x_{i,t} - \hat{x}_{i,t-1} \rvert > \Theta \\
0,                         & \text{otherwise}
\end{cases}
\end{equation}

The same update rules are applied to the hidden state \(h_{t}\). Eqs.~\eqref{eq:eq1}-\eqref{eq:x_del_it} summarize the delta principle in \cite{neil2016delta}.

Until now, this principle has been applied only to the inference process, i.e., the forward propagation phase.
Here we show that the delta sparsity can also be exploited in the backward propagation during the training process. 

\subsubsection{Intuition}
In the gradient-descent approach, we adjust the network weights to reduce the loss, and the gradient of a state vector depends on: a) how much the vector contributes to the network output; b) how much the output deviates from the ground truth.
The changes of inactivated neurons are discarded in forward propagation (Fig.~\ref{fig:dn_fp}) and make no contribution to the output, so their gradients are not needed. Therefore, we only need to propagate the errors of the changes of activated neurons, and calculate the weight gradients of the corresponding connections (Fig.~\ref{fig:dn_bp}). 
Detailed mathematical proofs for vanilla RNN, \tip{gru}~\cite{Cho2014}, and \tip{lstm}~\cite{lstm_hoch97} are given in Appendix A. 

\subsubsection{Proof Outline for Sparse \tip{bptt} in Vanilla Delta RNNs}
To allow later skipping the unchanging activations, we store a binary mask vector \({m}_{t}\) during forward propagation:
\begin{equation} \label{eq:mask}
{m}_{i,t} =
\begin{cases}
1, & \text{if } \lvert x_{i,t} - \hat{x}_{i,t-1} \rvert > \Theta \\
0, & \text{otherwise}
\end{cases}
\end{equation}
which indicates the neurons that are activated at the \(t\)-th time step during forward propagation. Then Eq.~\eqref{eq:x_del_it} becomes:
\begin{equation} \label{eq:x_del_t_m}
{\Delta x}_{t} = {m}_{t} \odot (x_{t} - \hat{x}_{t-1})
\end{equation}

In backward propagation, as shown in Fig.~\ref{fig:cg_bp},
we only compute the partial derivative of the cost \(C\) with respect to the change \({\Delta x}_{t}\) of the activated neurons:
\begin{equation} \label{eq:dc_ddx}
\pdv{C}{{\Delta x}_{t}} = \left( W_{x}^{\intercal} \frac{\partial C}{\partial M_{t}} \right) \odot {m}_{t}
\end{equation}
and calculate the weight gradients of those activated neurons:
\begin{equation} \label{eq:dc_dWx}
\frac{\partial C}{\partial W_{x}} = \sum_{t=1}^{T} \frac{\partial C}{\partial M_{t}} {\Delta x}_{t}^{\intercal} = \sum_{t=1}^{T} \frac{\partial C}{\partial M_{t} }  ({m}_{t} \odot (x_{t} - \hat{x}_{t-1}))^{T}
\end{equation}
where \(C=\sum_{t=1}^{T} {L_{t}}\) is the loss function values \(L_{t}\) summed across all time steps, \(\odot\) denotes the element-wise multiplication, and \(T\) is the total number of time steps. The gradients for \(\Delta h_{t}\) can be derived similarly.
More specifically, 
the vector \(\frac{\partial C}{\partial {\Delta x}_{t}}\) is not sparse, but the gradients of the inactivated neurons will be zeroed out during backward propagation due to the non-differentiability of Eq.~\eqref{eq:x_del_it} in the below-threshold case. So those values in \(\frac{\partial C}{\partial {\Delta x}_{t}}\) are not needed, and we can treat them as zeros using $m_t$ and skip their computations.

According to the formulations in Appendix A, the sparse versions of the backward propagation equations (Eqs.~\ref{eq:dc_ddx} and~\ref{eq:dc_dWx}) are equivalent to the dense versions for \tip{dn}, i.e. they result in exactly the same weight changes. Therefore, exploiting temporal sparsity in backward propagation will not cause extra accuracy loss when the delta threshold has already been applied in forward propagation.

Fig.~\ref{fig:3_sparse_mxv} illustrates how the \tip{dn} uses temporal sparsity in the \tip{mxv} operations arising from \(\Delta x_{t}\) in the forward (Eq.~\ref{eq:m_t}) and backward (Eqs.~\ref{eq:dc_ddx} and~\ref{eq:dc_dWx}) computations. We can store the indices of activated neurons at each timestep as a binary mask \(m_t\) or a \tip{nzil} during forward propagation. Since \(m_t\) can be applied to Eqs.~\eqref{eq:x_del_t_m}-\eqref{eq:dc_dWx}, the \tip{mxv} operations in backward propagation can share exactly the same sparsity. This sparsity allows for skipping entire columns of the weight matrices, thus the sparsity pattern is hardware-friendly.


\subsection{Theoretical Reduction in Computations and Memory Accesses}
\label{subsec:reduction_calc}


\textbf{Proposition 1.}  \textit{In training a vanilla RNN layer (described in Eq.~\ref{eq:eq1} and~\ref{eq:eq2}), the computational cost for calculating the gradients of weights $W_{x,t}$ or $W_{h,t}$ during each BPTT time step decreases linearly with the sparsity of delta input $\Delta x_t$ or delta states $\Delta h_t$ respectively. The total computation cost for the gradients of $W_{x}$ or $W_{h}$ with BPTT is the sum of terms proportional to the sparsity of $\Delta x_t$ or $\Delta h_t$ at each time step.}

The intuition for Proposition 1 is provided in Fig.~\ref{fig:cg_dn} and ~\ref{fig:3_sparse_mxv}.
The computational complexities of most commonly used optimizers (SGD, Adam~\cite{KingBa15}) are linear to the complexity of computing the gradients of cost function w.r.t. the weights.

To compute the theoretical reduction of computation and memory access resulting from the sparse delta vectors, we define \(o_c\) to be the occupancy of a vector, i.e., the ratio of non-zero elements in a vector.
According to the calculations in~\cite{neil2016delta}, the computation and memory access costs for sparse \tip{mxv} in forward propagation is reduced to:
\begin{equation}
C_{\text{sparse}}/C_{\text{dense}} \approx (o_c \cdot n^2) / n^2 = o_c
\end{equation}
where \(n\) is the size of the delta vector.
For example, for a \({\Delta x}_{t}\) sparsity of 90\%, the occupancy \(o_c\) is 10\%, the computation and memory access costs are reduced to 10\%, thus the theoretical speedup factor is \(1/o_c\) = 10X.

For the backward propagation in training, the computation cost for each timestep in Eq.~\eqref{eq:dc_dWx} is
\begin{equation*}
C_{\text{comp,sparse}} = o_c \cdot n^2
\end{equation*}
for a weight matrix \(W_x\) of size \(n \times n\) and a sparse \({\Delta x}_t\) vector of occupancy \(o_c\).
The memory access cost is
\begin{equation*}
C_{\text{mem,sparse}} = o_c \cdot n^2 + 3n
\end{equation*}
consisting of fetching \(o_c \cdot n^2\) weights for \(W_x\), reading \(n\) values for \({\Delta x}_t\) and \(n\) values for the mask \(m_t\), and writing \(n\) values for \(\pdv{C}{W_{x,t}}\). The costs for Eq.~\eqref{eq:dc_ddx} can be derived analogously. Therefore, the costs of matrix operations in backward propagation are also reduced to:
\begin{equation*}
C_{\text{sparse}}/C_{\text{dense}} \approx (o_c \cdot n^2) / n^2 = o_c
\end{equation*}
Because the sparsity in Eqs.~(\ref{eq:dc_ddx}) and (\ref{eq:dc_dWx}) is exactly the same as in Eq.~\eqref{eq:x_del_t_m}, the reduction factor in computation and memory access are also \((1 - o_c)\) equaling the sparsity of \({\Delta x}_{t}\). This illustrates the nice property of Delta Networks, that is, once we induce temporal sparsity in forward propagation, it can be exploited in all the three \tips{mxv} in both forward and backward propagation with nearly the same efficiency.

The sparse \tip{dn} training method requires marginal additional memory storage. Each timestep in forward propagation we store a binary mask \(m_t\) for \({\Delta x}\) or \({\Delta h}\) which only takes up \(n\) bits in the memory for a delta vector of size \(n\).

\section{Experiments}
\label{sec:experiments}

In this section, we first compare the accuracy and sparsity of original RNN models and Delta RNN models to verify the mathematical correctness of the sparse \tip{bptt} training method.
Next, we evaluate the performance of Delta RNNs on speech tasks, for both batch-32 training from scratch and batch-1 incremental learning. Finally we establish the benchmark of a custom hardware accelerator designed for training Delta RNNs.
For software experiments we implement Delta RNNs in Pytorch using custom functions for forward and backward propagation. Software experiments are conducted on a GTX 2080 Ti GPU. The hardware accelerator is implemented using \tip{hdl} and benchmarked in the Vivado simulator.

\subsection{Dense and Sparse BPTT Experiments}
\label{subsec:bptt_exp}

\begin{table*}[t]
  \centering
  \begin{threeparttable}
  \begin{tabular}{ccccccrrccc}
    \toprule
\multirow{2}{*}{Model} & \multirow{2}{*}{\(\Theta\)} &  & \multicolumn{2}{c}{MxV\tnote{*}} & \multirow{2}{*}{\begin{tabular}[c]{@{}c@{}}Accuracy\\ (\%)\end{tabular}} & \multicolumn{2}{c}{MACs (K)} & \multirow{2}{*}{} & \multicolumn{2}{c}{Sparsity (\%)} \\ \cline{4-5} \cline{7-8} \cline{10-11} 
 &  &  & FP & BP &  & \multicolumn{1}{c}{FP} & \multicolumn{1}{c}{BP} &  & FP & BP \\
    \midrule
LSTM        & -      &  & D   & D     & 93.1    & 73.7    & 147.5    &  & -       & -       \\
\textbf{Delta LSTM}  & 0.1    &  & \textbf{Sp}  & \textbf{D}     & \textbf{92.5}    & 12.2    & 147.5    &  & \textbf{83.4}    & 0       \\
\textbf{Delta LSTM}  & 0.1    &  & \textbf{Sp}  & \textbf{Sp}    & \textbf{92.5}    & 12.2    & 24.5     &  & \textbf{83.4}    & \textbf{83.4}    \\
Delta LSTM  & 0.2    &  & Sp  & D     & 91.0    & 5.2     & 147.5    &  & 92.9    & 0       \\
Delta LSTM  & 0.2    &  & Sp  & Sp    & 91.0    & 5.2     & 10.5     &  & 92.9    & 92.9    \\
\midrule
GRU         & -      &  & D   & D     & 93.6    & 55.3    & 110.6    &  & -       & -       \\
Delta GRU   & 0.1    &  & Sp  & D     & 92.7    & 13.1    & 110.6    &  & 76.3    & 0       \\
Delta GRU   & 0.1    &  & Sp  & Sp    & 92.7    & 13.1    & 26.2     &  & 76.3    & 76.3    \\
Delta GRU   & 0.2    &  & Sp  & D     & 91.6    & 6.6     & 110.6    &  & 88.0    & 0       \\
Delta GRU   & 0.2    &  & Sp  & Sp    & 91.6    & 6.6     & 13.3     &  & 88.0    & 88.0    \\
    \bottomrule
  \end{tabular}
  \begin{tablenotes}
    \item[*] Whether the \tips{mxv} operations are sparse. “Sp” means sparse and “D” is dense.
  \end{tablenotes}
  \end{threeparttable}
  \caption{Test accuracy of RNN models with dense and sparse training methods on GSCD.
  Network size is 16-128H-12 for all models. The numbers of parameters are 73.7k and 55.3k for LSTM and GRU models respectively. For FP/BP MACs, only \tip{mac} operations per timestep in RNN layers are taken into account. Boldfaced items highlight a group of Delta LSTM models trained with dense/sparse \tip{bptt} which results in identical accuracy and sparsity.}
  \label{tab:exp_gscd}
\end{table*}

The mathematical correctness of the sparse version of \tip{bptt} can be verified with random input data or real world datasets. In this experiment we use the \tip{gscd} v2~\cite{warden2018speech}, which contains 105,829 utterances of 35 English words. We split the dataset into train/validation/test sets with the ratio 8:1:1. 
The network model is a 16I-128H-12 (normal or Delta) LSTM/GRU model with a fully-connected layer for classification. Each model is trained for 40 epochs with learning rate 1e-3 and batch size 32. Two different fixed \(\Theta\) thresholds are tested for Delta LSTM/GRU models. We use the ADAM optimizer and weight decay coefficient of 1e-2. Test results are averaged over 5 runs with different random seeds. The classification result is obtained at the end of the presentation of each utterance.

\subsubsection{Results}
Table ~\ref{tab:exp_gscd} shows the results. Row 1 of each section (LSTM/GRU) shows the accuracy and cost of original LSTM. The other rows show the Delta LSTM results. “BP/D” denotes training with original dense \tip{bptt} by automatic differentiation in Pytorch, and “BP/Sp” is the proposed sparse \tip{bptt} method implemented by a custom backward function that performs zero-skipping with the mask vector in Eq.~\eqref{eq:mask}. Columns 3 and 4 show that when temporal sparsity is created in the forward pass (FP/Sp), the test accuracy is identical for dense (BP/D) and sparse (BP/Sp) \tip{bptt}. These results illustrate the mathematical equivalence of the masked \tip{bptt} equations with the original equations 
In addition, the sparsity is also identical for both forward and backward propagation, so we can save the same proportion of \tip{mxv} operations in both passes. We can draw the same conclusion from other rows in the table with different Delta thresholds and network architectures (LSTM and GRU).
\subsection{Spoken Language Training from Scratch Experiments}
\label{subsec:train_exp}


\begin{figure}[t]
    \centering
    \includegraphics[width=1\columnwidth]{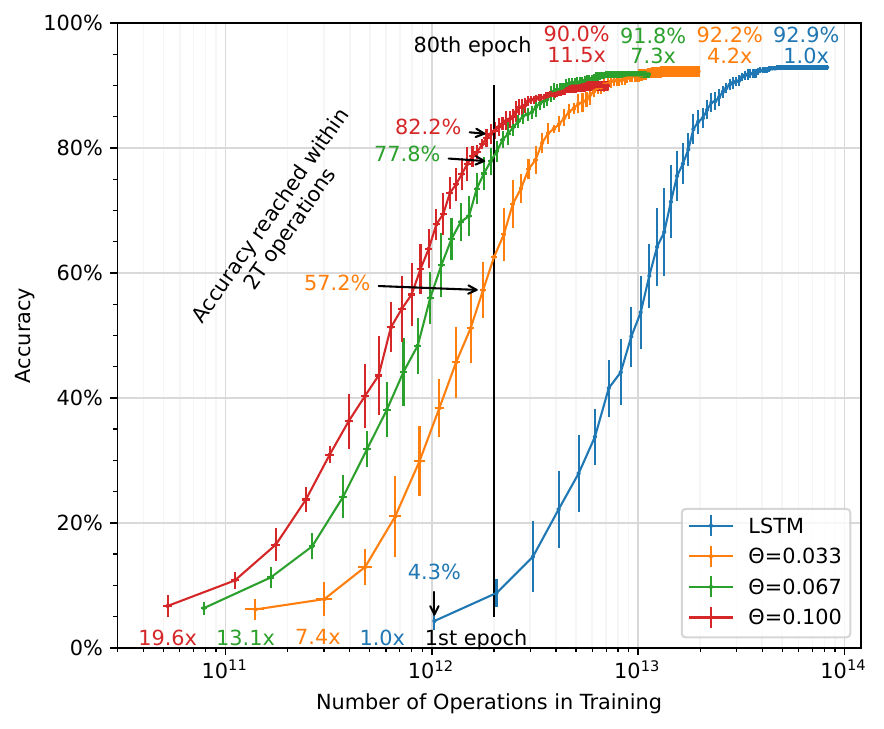}
    \caption{\tip{fscd} test accuracy vs number of training operations for LSTMs and Delta LSTMs. For each point, the x-coordinate is the number of \tip{mxv} operations performed up to this epoch, and the y-coordinate is the test accuracy at this epoch. The top and bottom data labels show the best accuracy achieved within 80 epochs and the reduction factor in operations compared with LSTM.}
    \label{fig:exp_fscd_acc_op}
\end{figure}


We next evaluated the accuracy and cost of training Delta RNNs for a speech task on the edge.
We use the \tip{fscd}~\cite{Lugosch2019} to evaluate the accuracy and cost of the Delta RNNs on \tip{slu} tasks. The dataset contains 30,043 utterances from 97 speakers for controlling smart-home appliances or virtual assistants. It has 248 \tip{slu} phrases mapping to 31 intents with three slots: action, object, and location, e.g. “Make it softer” is the same as “Turn down the volume”. The network model is a two-layer LSTM or Delta LSTM that both have 64 neurons, followed by one fully connected layer for classification. The model is trained for 80 epochs with learning rate 1e-3 and batch size 32. We use cosine annealing scheduler, ADAM optimizer, and weight decay coefficient of 1e-2. Test results are averaged over 5 runs with different random seeds.

\subsubsection{Results}
Fig.~\ref{fig:exp_fscd_acc_op} compares the classification accuracy versus training cost of a standard LSTM against Delta LSTMs with various delta thresholds. 
When the delta threshold \(\Theta\) increases, the number of operations needed to train the model to a given accuracy decreases dramatically, but the accuracy of Delta LSTM only slightly decreases.
After a set 80 epochs of training, the Delta LSTM with \(\Theta\)=0.067 (green curve) requires 7.3X fewer training operations than the LSTM with only 
a factor of 1.155X increase in error rate.
When computing resources are limited, the Delta Network can offer accurate training with an acceptable accuracy loss.

\subsection{Incremental Keyword Learning Experiments}
\label{subsec:IL_exp}


\begin{table*}[t]
  \centering
  \begin{tabular}{ccccccc}
    \toprule
\begin{tabular}[c]{@{}c@{}}CIL\\ setting\end{tabular} & Model & \begin{tabular}[c]{@{}c@{}}Batch\\ size\end{tabular} & \begin{tabular}[c]{@{}c@{}}Accuracy\\ (\%)\end{tabular} & \begin{tabular}[c]{@{}c@{}}Sparsity\\ (\%)\end{tabular} & \multicolumn{1}{c}{\begin{tabular}[c]{@{}c@{}}MACs\\ (K)\end{tabular}} & \begin{tabular}[c]{@{}c@{}}DRAM W accesses\\ (K words)\end{tabular} \\ \midrule
\multirow{2}{*}{35}      & LSTM                & 32         & 90.8 & -             & 221.2                & 221.2                \\
                         & Delta LSTM          & 32         & 90.5 & 81.7          & 40.4                 & 221.2                \\ \midrule
\multirow{2}{*}{20+3x5}  & LSTM                & 1          & 82.3 & -             & 221.2                & 221.2                \\
                         & \textbf{Delta LSTM} & \textbf{1} & 80.3 & \textbf{79.8} & \textbf{44.7}        & \textbf{44.7}        \\ \midrule
\multirow{2}{*}{20+1x15} & LSTM                & 1          & 76.6 & -             & 221.2                & 221.2                \\
                         & \textbf{Delta LSTM} & \textbf{1} & 74.8 & \textbf{79.8} & \textbf{44.7}        & \textbf{44.7}        \\ \bottomrule
  \end{tabular}
  \caption{Test accuracy, sparsity, and number of operations for \tip{cil} with LSTM models and Delta LSTM models (\(\Theta\)=0.1) on GSCD. For all models the network size is 16-128H-12 with 73.7k parameters.}
  \label{tab:exp_gscd_il}
\end{table*}

An attractive application of \tips{dn} is for online incremental training, where new labeled data become available in the field to an edge device and must be incorporated into the RNN to personalize or improve accuracy. 
To evaluate the performance of Delta RNNs on \tip{cil} tasks, we use GSCD v2~\cite{warden2018speech}, a dataset frequently used for benchmarking ASIC and FPGA keyword spotting implementations~\cite{shan2020510,giraldo2021efficient}.
The dataset is divided into train/test sets with the ratio 8:2.
We employ iCaRL~\cite{rebuffi2017icarl} as the incremental learning algorithm and use its evaluation method. The test accuracy is averaged over 5 runs with random permutation of classes, so for each run the order of classes learned by the model is different.
The network model is a one-layer LSTM or Delta LSTM that has 128 neurons, followed by one fully-connected layer for classification. We pretrain the model to learn 20 classes in 20 epochs with learning rate 1e-3 for batch-32 and learning rate 1e-4 for batch-1. Then we retrain the model to learn several new classes in 20 epochs with the same learning rates for each incremental learning task, during which the model only has access to a limited set of 2000 exemplars of the previously learned classes, until the model finishes all tasks and learns all 35 classes.
We use ADAM optimizer and weight decay coefficient of 1e-2.


\subsubsection{Results}
Table ~\ref{tab:exp_gscd_il} shows the results. The first column is the \tip{cil} setting, and we group the experiment results according to 3 different settings. The first setting “35” is the baseline where the network model learns all 35 classes directly. “20+3x5” means pretraining the model to learn 20 classes and then retraining the model to learn 3 additional classes each step for 5 times.
The fourth column shows the final accuracy after learning all 35 classes. The second last column shows the number of \tip{mac} operations per timestep in the LSTM layer. The last column shows the number of DRAM accesses for weights or weight gradients per timestep per batch in the LSTM layer.

It is clear from the table that the Delta LSTM models have \(\sim\)80\% sparsity and can save this proportion of \tip{mac} operations during training. Moreover, because the weight columns of inactivated neurons are not used during training and the memory access of them can be skipped as shown in Fig.~\ref{fig:3_sparse_mxv}, we can save \(\sim\)80\% memory access in the batch-1 training of Delta RNNs.
This significantly reduces energy consumption during training, because memory access consumes at least 10X more energy than arithmetic operations with the same bit width~\cite{horowitz20141}.
In an online setting, updating the weights using a batch size of 1 is natural and also desirable if the on-chip memory resources are too limited.

\subsection{Hardware Simulation of Delta Training Accelerator}
\label{subsec:acc_sim}

\begin{figure}[t]
    \centering
    \begin{subfigure}[c]{1\columnwidth}
        \centering
        \includegraphics[width=0.6\textwidth]{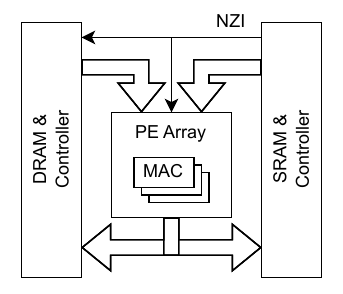}
        \caption{Accelerator block diagram.}
        \label{fig:sim_bd}
    \end{subfigure}
    \begin{subfigure}[c]{1\columnwidth}
        \centering
        \includegraphics[width=1\textwidth]{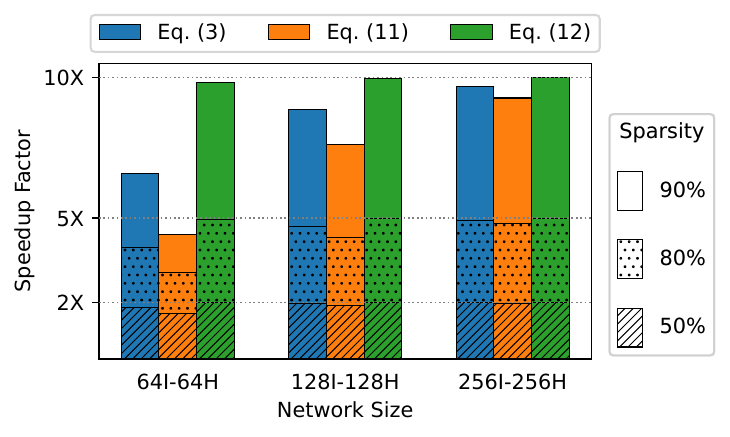}
        \caption{Speedup measured by RTL simulation of the accelerator.}
        \label{fig:sim_speedup}
    \end{subfigure}
    \caption{System architecture and Vivado simulation results of hardware accelerator for training Delta RNNs.}
    \label{fig:sim}
\end{figure}

To demonstrate the speedup of Delta RNNs, we evaluated the potential performance of a hardware training accelerator, through the clock-cycle accurate \tip{rtl} simulation of the design frequently used for logic design on FPGAs and custom ASICs.
The \tip{rtl} simulation closely emulates the actual hardware latency and logic costs of a custom silicon implementation.
The accelerator (Fig.~\ref{fig:sim_bd}) mainly consists of a \tip{pe} array where each \tip{pe} can perform a \tip{mac} operation in parallel at each clock cycle.
To efficiently exploit the temporal sparsity of Delta RNNs, the accelerator stores sparse activation vector data in its \tip{sram} in the format of \tip{nzil} and \tip{nzvl}, which enables the accelerator to skip computation (and \tip{dram} memory access of entire weight columns if the batch size is 1) for zero elements in delta vectors (Fig.~\ref{fig:3_sparse_mxv}) \cite{GaoDeltaRNN2018}.
Dynamically skipping weight columns ideally matches properties of burst mode \tip{dram} memory access, where addressing \tip{dram} columns is slow but reading them out is fast.

\subsubsection{Accelerator Computation Flow}
\begin{itemize}
\item Eq.~\eqref{eq:m_t}, FP: The accelerator reads the \tip{nzil} of input \({\Delta x}_t\), which is a list containing indices of activated neurons at \(t\)-th timestep, and fetches the corresponding weight columns from \tip{dram}, then multiply it with the \tips{nzv} of \({\Delta x}_t\).
\item Eq.~\eqref{eq:dc_ddx}, BP input gradient: The same weight columns are fetched using the \({\Delta x}_t\) \tip{nzil}, and they are multiplied with the gradient of pre-activation \(\frac{\partial C}{\partial M_{t}}\) to produce the gradients of activated neurons in \({\Delta x}_t\).
\item Eq.~\eqref{eq:dc_dWx}, BP weight gradient: The accelerator fetches the gradient of pre-activation \(\frac{\partial C}{\partial M_{t}}\) and multiply it with the \tips{nzv} of \({\Delta x}_t\), producing the gradient of weight columns \(\frac{\partial C}{\partial W_{x}}\) for activated neurons. The \tip{nzil} is used for output indexing in this step.
\end{itemize}
The same operations are performed for hidden states \(\Delta h\).
The accelerator process input samples one by one, i.e., the batch size of training is 1. This way, the accelerator can skip both the \tip{dram} access of weight columns of inactivated neurons and the computation for them.

\subsubsection{Experiment Setup}
Fig.~\ref{fig:sim_bd} shows the architecture of the accelerator that we instantiated with 16 \tips{pe}. It computes Eqs.~(\ref{eq:m_t}), (\ref{eq:dc_ddx}) and (\ref{eq:dc_dWx}), which are the three \tips{mxv} of training a Delta RNN. We tested three different network sizes: 64, 128 and 256. For each network, the input size equals the hidden layer size, which also equals the number of timesteps of input data. Random input data of different sparsities (50\%, 80\%, and 90\%) are generated to evaluate the performance of the accelerator. 
We measure the computation time in clock cycles for each \tip{mxv} equation, starting from the time when the input data is loaded into memory, to the time when PE array outputs the last data.

If there is no memory latency or communication overhead, and \tips{pe} are fully utilized, the computation time (in clock cycles) for a dense \tip{mxv} during RNN training is:
\begin{equation}
T_\text{dense} = \frac {N_{l} * N_{l-1} * T_{s}} {P}
\end{equation}
where \(N_l\), \(T_s\), and \(P\) denote the size of the \(l\)-th layer, the length of the \(s\)-th input sequence, and the number of \tips{pe} in the accelerator respectively.
The speedup factor is given by:
\begin{equation}
F_\text{speedup} = \frac {T_\text{dense}} {T_\text{measured}}
\end{equation}
where \(T_\text{measured}\) is the measured computation time in simulation. We compare the ideal result with the realistic \tip{rtl} simulation of the accelerator.

\subsubsection{Experiment Results}
Fig.~\ref{fig:sim_speedup} shows the speedup factors measured in the RTL simulations for different network sizes and activation sparsities. For input data of 50\%, 80\%, and 90\% sparsity, the 256I-256H Delta RNN nearly achieves 2X, 5X, and 10X training speedup, which are the theoretical speedup factors calculated in Section ~\ref{subsec:reduction_calc}. For smaller networks such as 64I-64H, the speedup factors for Eqs.~(\ref{eq:m_t}) and (\ref{eq:dc_ddx}) are lower, especially when the sparsity is high. This is because the two equations are calculated timestep by timestep, and the computation time of each timestep is short for small network or highly sparse data, making the overhead of communication and computation relatively more apparent in this case. The weight gradient calculation for all timesteps is executed at the end of \tip{bptt}, so the speedup factors of Eq.~\eqref{eq:dc_dWx} are close to the theoretical values for all tested cases.

\subsubsection{Summary}
In a custom training Delta RNN hardware accelerator using the \tip{nzil}-\tip{nzvl} data format, we can see a significant boost in the speedup factor with high activation sparsity.
The Delta RNN training accelerator would improve RNN incremental batch-1 training by a factor of about 5-10X compared to a dense RNN training accelerator.

\section{Related Works}
\label{sec:related_works}


The computational efficiency of neural networks can be improved by introducing sparsity into the networks.
\cite{o2016sigma} proposed an approach similar to Delta Network for CNNs to reduce the inference cost but \cite{aimar2019,neil2016delta} showed that using \tip{dn} requires doubled activation memory access because it must be read to check if it has changed, and then written, which ends up doubling the CNN inference cost because it is dominated by activation memory. By contrast, using the \tip{dn} on RNNs is beneficial because the fully-connected RNNs are weight-memory bounded, and the energy saving brought by temporal sparsity is much larger for RNNs.
\cite{gao2022spartus} and \cite{hunter2022two} exploit both \tip{dn} activation sparsity and weight sparsity to achieve impressive inference performance on hardware. 
Another method that can create sparsity in neural networks is conditional computation, or skipping operations.
Zoneout \cite{krueger2017zoneout} randomly selects whether to carry forward the previous hidden state or update it with the current hidden state during training. It aims to prevent overfitting in RNNs and provides only limited sparsity.

Skip RNN \cite{campos2017skip} uses a gating mechanism to learn whether to update or skip hidden states at certain timesteps, which is trained to optimize the balance between computational efficiency and accuracy. The skipping is applied to the whole RNN layers, rather than element-wise on activation vectors as in Delta RNNs.

EGRU \cite{subramoneyefficient} uses event-based communication between RNN neurons, resulting in sparse activations and a sparse gradient update.  While this method resembles \tip{dn}, its activation sparsity is different from the temporal sparsity of our work, and its sparsity in backward propagation is smaller by a factor of 1.5X to 10X than in forward propagation, unlike our Delta RNNs where the sparsity is identical for FP and BP. This asymmetry in forward and backward sparsity results from their surrogate gradient function which is non-zero within a certain range around the threshold to allow the errors near that point to pass through. Their surrogate gradient function makes the activation function differentiable, resulting in more accurate inference, but it can greatly decrease the sparsity in the backward pass. 

\cite{perez-nieves2021sparse} also touches on sparse \tip{bptt} but uses \tips{snn}. In a similar spirit, the authors show that computations can be saved by calculating the gradients only for active neurons (i.e. neurons producing a spike as defined by a threshold). However, the set of active neurons in backward propagation can be different from those in forward propagation. In contrast, the temporal sparsity in our work is identical in both the forward and backward propagation, thus the mask vectors computed in the forward pass can be directly reused in the backward pass. 
Their asynchronous \tip{snn} cannot efficiently use \tip{dram} for weight memory because the weight memory accesses are unpredictable.

\section{Conclusion}
\label{sec:conclusion}
Training RNNs with \tip{bptt} involves a huge number of arithmetic operations and especially memory accesses, 
which leads to inefficient deployment on edge platforms.
This paper shows that the temporal sparsity introduced in the Delta Network inference can also be applied during training leading to a sparse \tip{bptt} process.
The \tips{mxv} operations in \tip{bptt} can be significantly sped up by skipping the computation and propagation of gradients for inactivated neurons. Our experiments and digital hardware simulations demonstrate that the number of matrix multiplication operations in training RNNs can be reduced by 5-10X with marginal accuracy loss on speech tasks. Furthermore, the number of memory accesses can also be reduced by the same factor if training with a batch size of 1 on a hardware accelerator specifically designed for Delta RNNs, saving substantial energy consumption. Therefore, our proposed training method would be particularly useful for continuous online learning on resource-limited edge devices that can exploit self-supervised data, such as errors between predictions and measurements.

\newpage

\section*{Acknowledgments}
This project has received funding from Samsung Global Research “Neuromorphic Processor Project” (NPP), and the European Union’s Horizon 2020 research and innovation programme under grant agreement No 899287 for the project “Neural Active Visual Prosthetics for Restoring Function” (NeuraViPeR).

\bibliography{main}





\appendix
\newpage
\input{appendices}



\end{document}

%% file: acronyms.tex
\usepackage[draft]{pdfcomment} 
\usepackage[hyperfirst=false,acronym,sort=none,shortcuts,nopostdot,style=super,nonumberlist,toc,nogroupskip]{glossaries}
\usepackage[usenames,dvipsnames]{xcolor} 
\glsdisablehyper 

\newcommand{\tipcolor}{Sepia} 

\newcommand{\accolor}[1]{\textcolor{\tipcolor}{#1}}

\newacronymstyle{myacro}
{%
  \GlsUseAcrEntryDispStyle{long-short}%
}%
{%
  \GlsUseAcrStyleDefs{long-short}%
}

\setacronymstyle{myacro}

\newcommand*{\tip}[1]{
    \ifglsused{#1}{
      {\pdftooltip{\accolor{\glsentryshort{#1}}}{\glsentrydesc{#1}}}%
    }{%
      \gls{#1}
    }%
}%

\newcommand*{\tips}[1]{
    \ifglsused{#1}{
      {\pdftooltip{\accolor{\glsentryshortpl{#1}}}{\glsentrydescplural{#1}}}%
    }{%
      \glspl{#1}
    }%
}%

\newacronym{aae}{AAE}{Average Angular Error}
\newacronym{abmof}{ABMOF}{Adaptive Block Matching Optical Flow}
\newacronym{accmem}{ACC Mem}{Accumulation Memory}
\newacronym{accreg}{ACC REG}{Accumulation Register}
\newacronym{acp}{ACP}{Accelerator Coherency Port}
\newacronym{adc}{ADC}{Analog-to-Digital Converter}
\newacronym{add}{ADD}{adder}
\newacronym{admm}{ADMM}{Alternating Direction Method of Multipliers}
\newacronym{aee}{AEE}{Average Endpoint Error}
\newacronym{aer}{AER}{Address-event-representation}
\newacronym{afe}{AFE}{Analog Front-End}
\newacronym{ai}{AI}{Artificial Intellegence}
\newacronym{alm}{ALM}{Adaptive Logic Module}
\newacronym{ampro}{AMPRO}{Advanced Mechanical Bipedal Experimental Robotics Prosthesis}
\newacronym{am}{AM}{acoustic model}
\newacronym{ann}{ANN}{Artificial Neural Network}
\newacronym{aps}{APS}{Active Pixel Sensor}
\newacronym{ap}{AP}{Activation Pipeline}
\newacronym{asic}{ASIC}{Application Specific Integrated Circuit}
\newacronym{at}{AT}{Adder Trees}
\newacronym{axis}{AXIS}{AXI-Stream}
\newacronym{axi}{AXI}{Advanced eXtensible Interface}
\newacronym{bbb}{BBB}{Beaglebone Black}
\newacronym{bbs}{BBS}{Bank Balanced Sparsity}
\newacronym{blen}{BLEN}{Burst Length}
\newacronym{bnn}{BNN}{Binary Neural Network}
\newacronym{bpf}{BPF}{Band-Pass Filter}
\newacronym{bptt}{BPTT}{Backpropagation Through Time}
\newacronym{bram}{BRAM}{Block Random Access Memory}
\newacronym{br}{BR}{Balance Ratio}
\newacronym{cbcsc}{CBCSC}{Column-Balanced Compressed Sparse Column}
\newacronym{cbtd}{CBTD}{Column-Balanced Targeted Dropout}
\newacronym{ccd}{CCD}{Charge-coupled Device}
\newacronym{cfg}{CFG}{configuration}
\newacronym{cil}{CIL}{Class-Incremental Learning}
\newacronym{clk}{CLK}{clock}
\newacronym{cmos}{CMOS}{Complementary metal–oxide–semiconductor}
\newacronym{cnn}{CNN}{Convolutional Neural Network}
\newacronym{cnt}{CNT}{counter}
\newacronym{conv}{CONV}{convolutional}
\newacronym{cots}{COTS}{Commodity Off-The-Shelf}
\newacronym{cpld}{CPLD}{Complex Programmable Logic Device}
\newacronym{cpu}{CPU}{Central Processing Unit}
\newacronym{cp}{CP}{column pointer}
\newacronym{csc}{CSC}{Compressed Sparse Column}
\newacronym{csr}{CSR}{Compressed Sparse Row}
\newacronym{ctc}{CTC}{Connectionist Temporal Classification}
\newacronym{ctrl}{CTRL}{Controller}
\newacronym{cuda}{CUDA}{Compute Unified Device Architecture} 
\newacronym{cudnn}{cuDNN}{CUDA Deep Neural Network library}
\newacronym{cv}{CV}{Computer Vision}
\newacronym{dac}{DAC}{Digital-to-Analog Converter}
\newacronym{das}{DAS}{Dynamic Audio Sensor}
\newacronym{davis}{DAVIS}{Dynamic and Active pixel Vision Sensor}
\newacronym{dbe}{DBE}{Digital Back-End}
\newacronym{ddr}{DDR}{Double Data Rate}
\newacronym{dec}{DEC}{Decoder}
\newacronym{dev}{DEV}{development}
\newacronym{deltarnn}{DeltaRNN}{Delta Recurrent Neural Network}
\newacronym{deltagru}{DeltaGRU}{Delta Gated Recurrent Unit}
\newacronym{deltalstm}{DeltaLSTM}{Delta Long-Short Term Memory}
\newacronym{dl}{DL}{Deep Learning}
\newacronym{dma}{DMA}{Direct Memory Access}
\newacronym{dm}{DM}{Delta Memory}
\newacronym{dnn}{DNN}{Deep Neural Network}
\newacronym{dn}{DN}{Delta Network}
\newacronym{dpe}{DPE}{Delta Processing Element}
\newacronym{dram}{DRAM}{Dynamic Random Access Memory}
\newacronym{dsp}{DSP}{Digital Signal Processing}
\newacronym{du}{DU}{Delta Unit}
\newacronym{dvs}{DVS}{Dynamic Vision Sensor}
\newacronym{dt}{DT}{Delta Training}
\newacronym{edflow}{EDFLOW}{Event-driven Optical Flow}
\newacronym{eds}{EDS}{Event-Driven System}
\newacronym{eie}{EIE}{Efficient Inference Engine}
\newacronym{emmc}{eMMC}{Embedded Multi Media Card}
\newacronym{ese}{ESE}{Efficient Speech Recognition Engine}
\newacronym{fast}{FAST}{Features from Accelerated Segment Test}
\newacronym{fcl}{FCL}{Fully-Connected Layer}
\newacronym{fc}{FC}{Fully-Connected}
\newacronym{fex}{FEx}{Feature Extractor}
\newacronym{fft}{FFT}{Fast Fourier Transform}
\newacronym{fifo}{FIFO}{First-In First-Out}
\newacronym{fllatc}{FLL-ATC}{Frequency locked-loop analog-to-time converter}
\newacronym{fmllr}{fMLLR}{feature-space maximum likelihood linear regression}
\newacronym{fp16}{FP16}{16-bit floating-point number}
\newacronym{fp32}{FP32}{32-bit floating-point number}
\newacronym{fp8}{FP8}{8-bit floating-point number}
\newacronym{fpga}{FPGA}{Field Programmable Gate Array}
\newacronym{fps}{FPS}{Frame Per Second}
\newacronym{fp}{FP}{floating-point}
\newacronym{fscd}{FSCD}{Fluent Speech Commands Dataset}
\newacronym{fsm}{FSM}{Finite-State Machine}
\newacronym{fxp}{FXP}{fixed-point}
\newacronym{gb}{GB}{gigabytes}
\newacronym{gemm}{GEMM}{General Matrix Multiplication}
\newacronym{gpio}{GPIO}{General-Purpose Input/Output}
\newacronym{gpu}{GPU}{Graphics Processing Unit}
\newacronym{gp}{GP}{General Purpose}
\newacronym{gru}{GRU}{Gated Recurrent Unit}
\newacronym{gscd}{GSCD}{Google Speech Command Dataset}
\newacronym{gt}{GT}{Ground Truth}
\newacronym{hbm}{HBM}{High Bandwidth Memory}
\newacronym{hcgs}{HCGS}{Hierarchical Coarse-Grain Sparsity}
\newacronym{hdl}{HDL}{Hardware Description Language}
\newacronym{hdr}{HDR}{high dynamic range}
\newacronym{hls}{HLS}{High Level Synthesis}
\newacronym{hmc}{HMC}{Hybrid Memory Cube}
\newacronym{hmm}{HMM}{Hidden Markov Model}
\newacronym{hpe}{HPE}{Heterogeneous Processing Element}
\newacronym{hp}{HP}{High Performance}
\newacronym{ht}{HT}{High Throughput}
\newacronym{idx}{IDX}{index}
\newacronym{ieu}{IEU}{Input Encoding Unit}
\newacronym{iir}{IIR}{Infinite Impulse Response}
\newacronym{imc}{IMC}{In-Memory Computing}
\newacronym{imu}{IMU}{Inertial Measurement Unit}
\newacronym{int16}{INT16}{16-bit integer}
\newacronym{int8}{INT8}{8-bit integer}
\newacronym{int}{INT}{integer}
\newacronym{iot}{IoT}{Internet-of-Things}
\newacronym{io}{I/O}{Input/Output}
\newacronym{ipu}{IPU}{Input Processing Unit}
\newacronym{ip}{IP}{Intellectual Property}
\newacronym{isa}{ISA}{Instruction Set Architecture}
\newacronym{jtag}{JTAG}{Joint Test Action Group}
\newacronym{kb}{KB}{kilobytes}
\newacronym{kws}{KWS}{Keyword Spotting}
\newacronym{lasso}{LASSO}{least absolute shrinkage and selection operator}
\newacronym{ldo}{LDO}{Low-dropout}
\newacronym{lidx}{LIDX}{local index}
\newacronym{ll}{LL}{Low Latency}
\newacronym{lstm}{LSTM}{Long Short-Term Memory}
\newacronym{lutram}{LUTRAM}{Look-up Table based Random Access Memory}
\newacronym{lut}{LUT}{look-up table}
\newacronym[description={Multiply-Accumulate is the basic operation of signal processing and artificial neural networks. One MAC is 2 Op.}]{mac}{MAC}{Multiply-Accumulate}
\newacronym{mb}{MB}{megabytes}
\newacronym{mcu}{MCU}{Microcontroller Unit}
\newacronym{mfcc}{MFCC}{Mel Frequency Cepstral Coefficients}
\newacronym{mkl}{MKL}{Math Kernel Library}
\newacronym{mlp}{NLP}{Multi-Layer Perception}
\newacronym{mm2s}{MM2S}{Memory Mapped to Stream}
\newacronym{mmp}{MMP}{Mini-Module-Plus}
\newacronym{mul}{MUL}{multiplier}
\newacronym{mxspv}{MxSPV}{Matrix-Sparse Vector Multiplication}
\newacronym{mxv}{MxV}{Matrix-Vector Multiplication}
\newacronym{nas}{NAS}{Neural Architecture Search}
\newacronym{ncs}{NCS}{Neural Compute Stick}
\newacronym{nlp}{NLP}{Natural Language Processing}
\newacronym{norm}{Norm}{normalization}
\newacronym{nss}{NSS}{Nyquist-Sample System}
\newacronym{nzil}{NZIL}{Non-Zero Index List}
\newacronym{nzi}{NZI}{Nonzero Index}
\newacronym{nzvl}{NZVL}{Non-Zero Value List}
\newacronym{nzv}{NZV}{Nonzero Value}
\newacronym{ob}{OBUF}{Output Buffer}
\newacronym{onn}{ONN}{Optical Neural Network}
\newacronym{os}{OS}{Operating System}
\newacronym{pcb}{PCB}{Printed Circuit Board}
\newacronym{pcol}{pcol}{weight column pointers}
\newacronym{pc}{PC}{Personal Computer}
\newacronym{pd}{PD}{proportional–derivative}
\newacronym{per}{PER}{Phone Error Rate}
\newacronym{pe}{PE}{Processing Element}
\newacronym{pfm}{PFM}{Pulse-frequency-modulated}
\newacronym{pl}{PL}{Programmable Logic}
\newacronym{prm}{PRM}{Pixel Rendering Module}
\newacronym{ps}{PS}{Processing System}
\newacronym{pt}{PT}{Previous Time}
\newacronym{pwm}{PWM}{Pulse-Width-Modulated}
\newacronym{qspi}{QSPI}{Quad Serial Peripheral Interface}
\newacronym{ralut}{RALUT}{Range Addressable Lookup Table}
\newacronym{ram}{RAM}{Random Access Memory}
\newacronym{ratp}{RATP}{Recursive Adaptive Temporal Pooling}
\newacronym{rec}{Rec}{Rectified}
\newacronym{relu}{ReLU}{Rectified Linear Unit}
\newacronym{reram}{ReRAM}{Resistive random-access memory}
\newacronym{ridx}{RIDX}{relative index}
\newacronym{risc}{RISC}{Reduced Instruction Set Computer}
\newacronym{rmse}{RMSE}{Root-Mean-Square Error}
\newacronym{rnn}{RNN}{Recurrent Neural Network}
\newacronym{ros}{ROS}{Robot Operating System}
\newacronym{rtl}{RTL}{Register Transfer Level}
\newacronym{s2mm}{S2MM}{Stream to Memory Mapped}
\newacronym{sad}{SAD}{Sum of Absolute Differences}
\newacronym{sae}{SAE}{Surface of Active Events}
\newacronym{sdk}{SDK}{Software Development kit}
\newacronym{sd}{SD}{Secure Digital}
\newacronym{se}{SE}{Single-Ended}
\newacronym{sits}{SITS}{Speed Invariant Time Surface}
\newacronym{slu}{SLU}{Spoken Language Understanding}
\newacronym{smem}{SMEM}{State Memory}
\newacronym{sm}{SM}{Supplementary Material}
\newacronym{snn}{SNN}{Spiking Neural Network}
\newacronym{snr}{SNR}{Signal-to-Noise Ratio}
\newacronym{soc}{SoC}{System-on-a-Chip}
\newacronym{som}{SOM}{System-on-Module}
\newacronym{spi}{SPI}{Serial Peripheral Interface}
\newacronym{spmxspv}{SPMxSPV}{Sparse Matrix-Sparse Vector Multiplication}
\newacronym{spmxv}{SPMxV}{Sparse Matrix-Vector Multiplication}
\newacronym{sram}{SRAM}{Static Random Access Memory}
\newacronym{sro}{SRO}{Switched-Ring-Oscillator}
\newacronym{test}{TEST}{test}
\newacronym{tf}{TF}{Threshold Function}
\newacronym{tsmc}{TSMC}{Taiwan Semiconductor Manufacturing Company}
\newacronym{usb}{USB}{Universal Serial Bus}
\newacronym{vad}{VAD}{Voice Activity Detection}
\newacronym{val}{VAL}{Value}
\newacronym{wer}{WER}{Word Error Rate}
\newacronym{wfst}{WFST}{Weighted Finite State Transducer}
\newacronym{wl}{WL}{workload}
\newacronym{wmem}{WMEM}{Weight Memory}
\newacronym{xor}{XOR}{Exclusive-OR}

%% file: appendices.tex
\begin{appendices}
\newcommand*\tcircle[1]{%
  \raisebox{-0.5pt}{%
    \textcircled{\fontsize{7pt}{0}\fontfamily{phv}\selectfont #1}%
  }%
}

\section{Delta RNNs BPTT Formulations}

This section shows the derivation of \tip{bptt} equations of Delta RNNs.
First we show the proof of Proposition 1 in Section~\ref{subsec:reduction_calc}.
Next we derive the \tip{bptt} formulation of Delta RNNs~\cite{neil2016delta}, then extend it to the two variants of RNNs: \tip{gru}~\cite{Cho2014} and \tip{lstm}~\cite{lstm_hoch97}.

\subsection{Proof of Proposition 1}

In this subsection we give the proof of Proposition 1 in Section~\ref{subsec:reduction_calc} for the gradient of weight \(W_{x}\). Proof for the gradient of weight \(W_{h}\) is similar.

In a Delta RNN layer, the computation in the forward propagation phase is formulated as:
\begin{equation} \label{eq:M_xh_t}
    M_{t} = M_{x,t} + M_{h,t} + M_{t-1}
\end{equation}
\begin{equation} \label{eq:M_x_t}
    M_{x,t} = \bm{W_{x}} {\Delta x}_{t}
\end{equation}
\begin{equation} \label{eq:M_h_t}
    M_{h,t} = \bm{W_{h}} {\Delta h}_{t-1}
\end{equation}
\begin{equation} \label{eq:appx_x_hat_it}
    \hat{x}_{i,t} =
    \begin{cases}
        x_{i,t},         & \text{if } \lvert x_{i,t} - \hat{x}_{i,t-1} \rvert > \Theta \\
        \hat{x}_{i,t-1}, & \text{otherwise}
    \end{cases}
\end{equation}
\begin{equation} \label{eq:appx_x_del_it}
    {\Delta x}_{i,t} =
    \begin{cases}
        x_{i,t} - \hat{x}_{i,t-1}, & \text{if } \lvert x_{i,t} - \hat{x}_{i,t-1} \rvert > \Theta \\
        0,                         & \text{otherwise}
    \end{cases}
\end{equation}
In the backward propagation phase, the cost function is the sum of the loss function at every timestep:
\begin{equation} \label{eq:cost}
    C = \sum_{t=1}^{T} {L_t}
\end{equation}
Hidden states $h_t$ is a function of $M_t$ associated with any activation function and loss function $L_t$ is a function of $h_t$ and ground truth value at that time step.
The gradient of weight \(W_{x}\) is the partial derivative of the cost \(C\) w.r.t. \(W_{x}\):
\begin{equation} \label{eq:dc_dWx_sum}
    \pdv{C}{W_x} = \pdv{(\sum_{t=1}^{T} {L_t})}{W_x} = \sum_{t=1}^{T} {\pdv{L_t}{W_x}}
\end{equation}
For each timestep, according to the chain rule:
\begin{equation} \label{eq:Lt_Wx_1}
    \pdv{L_t}{W_x} = \pdv{L_t}{h_t} \pdv{h_t}{W_x} = \pdv{L_t}{h_t} \pdv{h_t}{M_t} \pdv{M_t}{W_x}
\end{equation}
From Eqs.~\eqref{eq:M_xh_t}-\eqref{eq:M_h_t} we can expand the last term:
\begin{equation}
\begin{aligned}
    \pdv{M_t}{W_x} & = \pdv{(W_{x} {\Delta x}_t + W_h {\Delta h}_{t-1} + M_{t-1})}{W_x} \\
                   & = {\Delta x}_t^{\intercal} + \pdv{M_{t-1}}{W_x}
\end{aligned}
\end{equation}
Using telescoping we get
\begin{equation}
    \pdv{M_t}{W_x} = {\Delta x}_t^{\intercal} + {\Delta x}_{t-1}^{\intercal} + \dots + {\Delta x}_{1}^{\intercal}
                   = \sum_{i=1}^{t} {{\Delta x}_i^{\intercal}}
\end{equation}
Then Eq.~\eqref{eq:Lt_Wx_1} becomes:
\begin{equation} \label{eq:Lt_Wx_2}
    \pdv{L_t}{W_x} = \pdv{L_t}{h_t} \pdv{h_t}{M_t} (\sum_{i=1}^{t} {{\Delta x}_i^{\intercal}})
\end{equation}

For simplicity we write
\begin{equation*}
    \pdv{L_t}{h_t} = L'(h_t), \qquad
    \pdv{h_t}{M_t} = h'(M_t)
\end{equation*}
Plug them into Eq.~\eqref{eq:dc_dWx_sum}, we get:
\begin{equation} \label{eq:dc_dWx_sum_2}
\begin{aligned}
    \pdv{C}{W_x} ={} & \sum_{t=1}^{T} {\pdv{L_t}{W_x}} \\
                 ={} & \sum_{t=1}^{T} {\bigl[ L'(h_t) h'(M_t) (\sum_{i=1}^{t} {{\Delta x}_i^{\intercal}}) \bigr]} \\
                 ={} & \underbrace{L'(h_1) h'(M_1) {\Delta x}_1^{\intercal}}_{t=1} \\
                     & + \underbrace{L'(h_2) h'(M_2) ({\Delta x}_1^{\intercal}+{\Delta x}_2^{\intercal})}_{t=2} \\
                     & + \dots + \underbrace{L'(h_T) h'(M_T) (\sum_{i=1}^{T} {{\Delta x}_i^{\intercal}})}_{t=T} \\
                 ={} & \sum_{t=1}^{T} {L'(h_t) h'(M_t) {\Delta x}_1^{\intercal}} \\
                     & + \sum_{t=2}^{T} {L'(h_t) h'(M_t) {\Delta x}_2^{\intercal}} \\
                     & + \dots + \sum_{t=T}^{T} {L'(h_t) h'(M_t) {\Delta x}_T^{\intercal}} \\
                 ={} & \sum_{t=1}^{T} \underbrace{\biggl[ \underbrace{\Bigl(\sum_{i=t}^{T} {L'(h_i) h'(M_i)}\Bigr)}_{\tcircle{1}} {\Delta x}_t^{\intercal} \biggr]}_{\tcircle{2}}
\end{aligned}
\end{equation}

{\large \textcircled{\small 1}} is a partial sum that has constant complexity in each timestep in \tip{bptt}.
{\large \textcircled{\small 2}} is a vector outer product performed at each timestep.
For a vector \({\Delta x}_t\) of size \(n\) and occupancy \(o_{c}\) (Section~\ref{subsec:reduction_calc}), the computation cost of {\large \textcircled{\small 2}} is
\begin{equation*}
    C_{\text{comp,sparse}} = o_{c} \cdot n^2
\end{equation*}
which decreases linearly with the sparsity of \({\Delta x}_t\).
The total cost of Eq.~\eqref{eq:dc_dWx_sum_2} is the sum of {\large \textcircled{\small 2}} across all timesteps, thus proportional to the sparsity of \({\Delta x}_t\) at each timestep.

\subsection{Delta RNN BPTT Formulation}
\label{subsec:bptt_deltarnn}



\begin{figure}[ht]
    \centering
    \begin{subfigure}[t]{.40\textwidth}
        \centering
        \includegraphics[width=\linewidth]{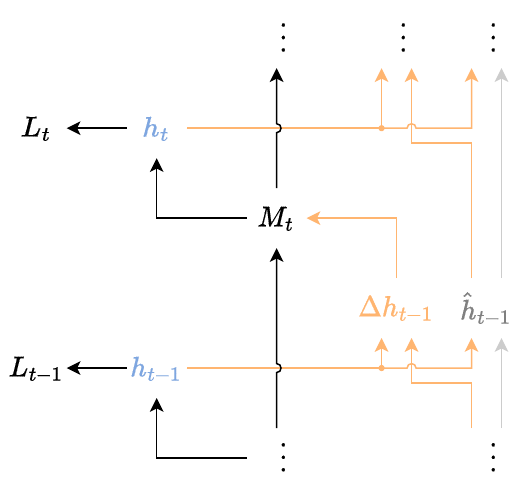}
        \caption{FP computation graph regarding \(h_t\).}
        \label{fig:cg_deltarnn_fp_h}
    \end{subfigure}
    \hfill
    \begin{subfigure}[t]{.40\textwidth}
        \centering
        \includegraphics[width=\linewidth]{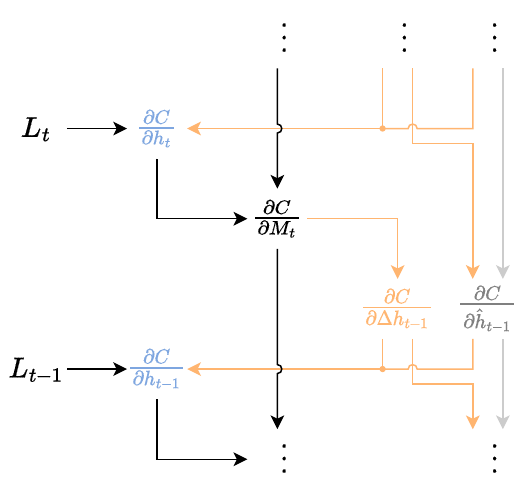}
        \caption{BP computation graph regarding \(h_t\).}
        \label{fig:cg_deltarnn_bp_h}
    \end{subfigure}
    \caption{Computation graphs of a Delta RNN layer. Weight matrices are omitted for clarity.}
    \label{fig:cg_deltarnn}
\end{figure}

The Forward Propagation (FP) equations of Delta RNNs~\cite{neil2016delta} are shown in Section~\ref{subsec:dn_form}. The FP computation graph is shown in Fig.~\ref{fig:cg_deltarnn_fp_h}. The lines in orange/gray color means that only values of activated/inactivated neurons are propagated along these paths. For clarity we only show the computation paths relevant to \(h_t\). The other parts regarding \(x_t\) can be drawn in the similar way.

In \tip{bptt}, the loss function \(L_{t}\) is computed for each timestep, which sums up to the cost:
\({C} = \sum_{t=1}^{T} {L_{t}}\).
The BP computation paths are reversed (Fig.~\ref{fig:cg_deltarnn_bp_h}) from the FP paths. The error is propagated backward through time, and the gradients are calculated along the paths according to the chain rule. The total gradient of a vertex in the graph is the sum of the gradients of all incoming paths.

The pre-activation memory \(M_t\) has two incoming connections in the BP computation graph, one from \(M_{t+1}\) and the other from \(h_t\). So the PD of \(M_t\) w.r.t. the cost \(C\) has two components:
\begin{equation} \label{eq:dc_dm}
\begin{split}
    \pdv{C}{M_{t}} & = \pdv{C}{h_{t}} \pdv{h_{t}}{M_{t}} + \pdv{C}{M_{t+1}} \overline{\pdv{M_{t+1}}{M_{t}}} \\
                   & = \pdv{C}{h_{t}} \tanh'(M_{t}) + \pdv{C}{M_{t+1}}
\end{split}
\end{equation}
We use an overline to denote the part of a Partial Derivative (PD) that only takes the direct connection into account. For example, \(\overline{\pdv{M_{t+1}}{M_{t}}}\) is the PD coming from the path \(M_{t+1} \rightarrow M_t\) directly, excluding any other path such as \(M_{t+1} \rightarrow {\Delta h}_t \rightarrow h_t \rightarrow M_t\).

Similarly, we can derive the other gradients in the computation graph as follows.
\begin{equation} \label{eq:dc_ddh}
    \pdv{C}{{\Delta h}_{t-1}} = \pdv{C}{M_{t}} \pdv{M_{t}}{{\Delta h}_{t-1}} = \bm{W_{h}^{\intercal}} \pdv{C}{M_{t}}
\end{equation}
\begin{equation} \label{eq:dc_dhh}
    \pdv{C}{{\hat h}_{t-1}} = \pdv{C}{{\hat h}_{t}} \pdv{{\hat h}_{t}}{{\hat h}_{t-1}} + \pdv{C}{{\Delta h}_{t}} \pdv{{\Delta h}_{t}}{{\hat h}_{t-1}}
\end{equation}
\begin{equation} \label{eq:dc_dh}
    \pdv{C}{h_{t}} = \pdv{C}{{\hat h}_{t}} \pdv{{\hat h}_{t}}{h_{t}} + \pdv{C}{{\Delta h}_{t}} \pdv{{\Delta h}_{t}}{h_{t}} + \pdv{L_{t}}{h_{t}}
\end{equation}
\begin{equation} \label{eq:dc_dWh}
    \bm{\frac{\partial C}{\partial W_h}} = \sum_{t=1}^{T} {\pdv{C}{M_{t}} \pdv{M_{t}}{W_{h}}} = \sum_{t=1}^{T} {\pdv{C}{M_{t}} {\Delta h}_{t-1}^{\intercal}}
\end{equation}
The pre-activation memory is initialized as \(M_0 = b_h\) in FP, so the gradient of the bias is \(\pdv{C}{b_h} = \pdv{C}{M_0}\).

The equations for gradients w.r.t. \(x_t\) are similar, except that there is no direct connection between \(x_t\) and the loss \(L_t\). So Eq.~\eqref{eq:dc_dh} can be modified for \(x_t\) as:
\begin{equation}
    \pdv{C}{x_{t}} = \pdv{C}{{\hat x}_{t}} \pdv{{\hat x}_{t}}{x_{t}} + \pdv{C}{{\Delta x}_{t}} \pdv{{\Delta x}_{t}}{x_{t}}
\end{equation}

To compute the first two terms in Eqs.~\eqref{eq:dc_dhh} and \eqref{eq:dc_dh}, we need to look back at the FP equations:
\begin{equation} \label{eq:h_hat_it}
    \hat{h}_{i,t} =
    \begin{cases}
        h_{i,t},         & \text{if } \lvert h_{i,t} - \hat{h}_{i,t-1} \rvert > \Theta \\
        \hat{h}_{i,t-1}, & \text{otherwise}
    \end{cases}
\end{equation}
\begin{equation} \label{eq:h_del_it}
    {\Delta h}_{i,t} =
    \begin{cases}
        h_{i,t} - \hat{h}_{i,t-1}, & \text{if } \lvert h_{i,t} - \hat{h}_{i,t-1} \rvert > \Theta \\
        0,                         & \text{otherwise}
    \end{cases}
\end{equation}

Eq.~\eqref{eq:h_hat_it} has two branches, thus the PDs of \(\hat{h}_{t}\) w.r.t. \({h}_{t}\) and \(\hat{h}_{t-1}\) also have two branches:
\begin{equation} \label{eq:dhh_dh_t}
    \pdv{\hat{h}_{i,t}}{{h}_{i,t}} =
    \begin{cases}
        1, & \text{if } \lvert h_{i,t} - \hat{h}_{i,t-1} \rvert > \Theta \\
        0, & \text{otherwise}
    \end{cases}
\end{equation}
\begin{equation} \label{eq:dhh_dhh_tm1}
    \pdv{\hat{h}_{i,t}}{\hat{h}_{i,t-1}} =
    \begin{cases}
        0, & \text{if } \lvert h_{i,t} - \hat{h}_{i,t-1} \rvert > \Theta \\
        1, & \text{otherwise}
    \end{cases}
\end{equation}
From Eq.~\eqref{eq:h_del_it} the PDs of \({\Delta h}_{t}\) w.r.t. \({h}_{t}\) and \(\hat{h}_{t-1}\) are:
\begin{equation} \label{eq:ddh_dh_t}
    \pdv{{\Delta h}_{i,t}}{{h}_{i,t}} =
    \begin{cases}
        1, & \text{if } \lvert h_{i,t} - \hat{h}_{i,t-1} \rvert > \Theta \\
        0, & \text{otherwise}
    \end{cases}
\end{equation}
\begin{equation} \label{eq:ddh_dhh_tm1}
    \pdv{{\Delta h}_{i,t}}{\hat{h}_{i,t-1}} =
    \begin{cases}
        -1, & \text{if } \lvert h_{i,t} - \hat{h}_{i,t-1} \rvert > \Theta \\
        0,  & \text{otherwise}
    \end{cases}
\end{equation}

For convenience we use a mask vector \({m}_{h,t}\) to memorize whether a neuron is activated during FP:
\begin{equation} \label{eq:mask_h}
    {m}_{h,i,t} =
    \begin{cases}
        1, & \text{if } \lvert h_{i,t} - \hat{h}_{i,t-1} \rvert > \Theta \\
        0, & \text{otherwise}
    \end{cases}
\end{equation}

Then Eqs.~\eqref{eq:h_hat_it} and \eqref{eq:h_del_it} can be rewritten as:
\begin{equation} \label{eq:h_hat}
    {\hat h}_{t} = h_{t} \odot {m}_{h,t} + {\hat h}_{t-1} \odot (1 - {m}_{h,t})
\end{equation}
\begin{equation} \label{eq:h_del}
    {\Delta h}_{t} = (h_{t} - {\hat h}_{t-1}) \odot {m}_{h,t}
\end{equation}
where \(\odot\) denotes the element-wise multiplication.

And the PDs in Eqs.~\eqref{eq:dhh_dh_t}-~\eqref{eq:ddh_dhh_tm1} can be rewritten as:
\begin{equation*}
    \pdv{{\hat h}_{t}}{h_{t}} = {m}_{h,t}
    \text{ ,} \qquad
    \pdv{{\hat h}_{t}}{{\hat h}_{t-1}} = 1 - {m}_{h,t}
\end{equation*}
\begin{equation*}
    \pdv{{\Delta h}_{t}}{h_{t}} = {m}_{h,t}
    \text{ ,} \qquad
    \pdv{{\Delta h}_{t}}{{\hat h}_{t-1}} = - {m}_{h,t}
\end{equation*}
Plug them into Eqs.~\eqref{eq:dc_dhh} and \eqref{eq:dc_dh}, we have:
\begin{equation} \label{eq:dc_dhh_masked}
    \pdv{C}{{\hat h}_{t-1}} = \pdv{C}{{\hat h}_{t}} \odot (1 - {m}_{h,t}) - \pdv{C}{{\Delta h}_{t}} \odot {m}_{h,t}
\end{equation}
\begin{equation} \label{eq:dc_dh_masked}
    \pdv{C}{h_{t}} = \pdv{C}{{\hat h}_{t}} \odot {m}_{h,t} + \pdv{C}{{\Delta h}_{t}} \odot {m}_{h,t} + \pdv{L_{t}}{h_{t}}
\end{equation}
The term \(\pdv{C}{{\Delta h}_{t}}\) is multiplied with the mask \({m}_{h,t}\) for all the equations relevant to it. In other words, only values at the indices of those neurons who are activated during FP are used in BP. So we only need to calculate the values in \(\pdv{C}{{\Delta h}_{t}}\) for activated neurons in Eq.~\eqref{eq:dc_ddh} and discard other values, just like we discard the changes of inactivated neurons \({\Delta h}_{t}\) in FP. In fact, Eq.~\eqref{eq:dc_ddh} can be calculated as:
\begin{equation} \label{eq:dc_ddh_masked}
    \pdv{C}{{\Delta h}_{t-1}} = \left( \bm{W_{h}^{\intercal}} \pdv{C}{M_{t}} \right) \odot {m}_{h,t-1}
\end{equation}
which would give exactly the same results Eqs.~\eqref{eq:dc_dhh_masked} and \eqref{eq:dc_dh_masked} and other equations in \tip{bptt}.

This can also be seen from Eq.~\eqref{eq:h_del_it}. \({\Delta h}_{t}\) is a constant 0 at the below-threshold branch which is not differentiable, so the PDs of \({\Delta h}_{t}\) w.r.t. \({h}_{t}\) and \(\hat{h}_{t-1}\) are 0 at this branch. According to the chain rule, any gradients of \({\Delta h}_{t}\) passing through this branch will be multiplied by these PDs, effectively zeroing out those gradients. Consequently, we only need the \({\Delta h}_{t}\) gradients for the above-threshold branch, which can be expressed by Eq.~\eqref{eq:dc_ddh_masked}.

Therefore, once the delta threshold \(\Theta\) is introduced in FP, the accuracy of the network may decrease due to the loss of some information, but the temporal sparsity created in \(\Delta x\) and \(\Delta h\) during FP can be leveraged in BP without causing any extra accuracy loss.

The main computations regarding the hidden state vector \(h_t\) in BP are the vector outer products in Eq.~\eqref{eq:dc_dWh} and the \tips{mxv} in Eq.~\eqref{eq:dc_ddh_masked}, both of them sparsified by the delta mask \({m}_{h,t}\). The gradients regarding the input vector \(x_t\) can be derived in a similar way, which would also results in sparse \tips{mxv} with the mask \({m}_{x,t}\).

In conclusion, in the \tip{bptt} phase of Delta RNNs, we only need to propagate the errors of the changes of activated neurons, and calculate the weight gradients of the corresponding connections. The three \tip{mxv} operations during the training of Delta RNNs, which are expressed by Eqs.~\eqref{eq:M_h_t}, \eqref{eq:dc_dWh}, and \eqref{eq:dc_ddh_masked}, share the same sparsity (Figs.~\ref{fig:3_sparse_mxv}). A large number of operations can be reduced by skipping computations of inactivated neurons in both forward and backward propagation using binary mask vectors or \tips{nzil}.

\subsection{Delta LSTM BPTT Formulation}

\begin{figure}[t!]
    \centering
    \begin{subfigure}[t]{.45\textwidth}
        \centering
        \includegraphics[width=\linewidth]{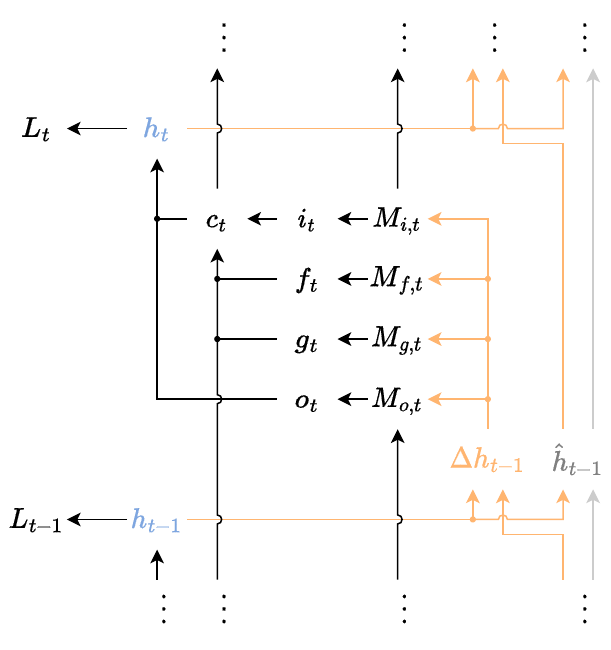}
        \caption{FP computation graph regarding \(h_t\).}
        \label{fig:cg_deltalstm_fp_h}
    \end{subfigure}
    \hfill
    \begin{subfigure}[t]{.45\textwidth}
        \centering
        \includegraphics[width=\linewidth]{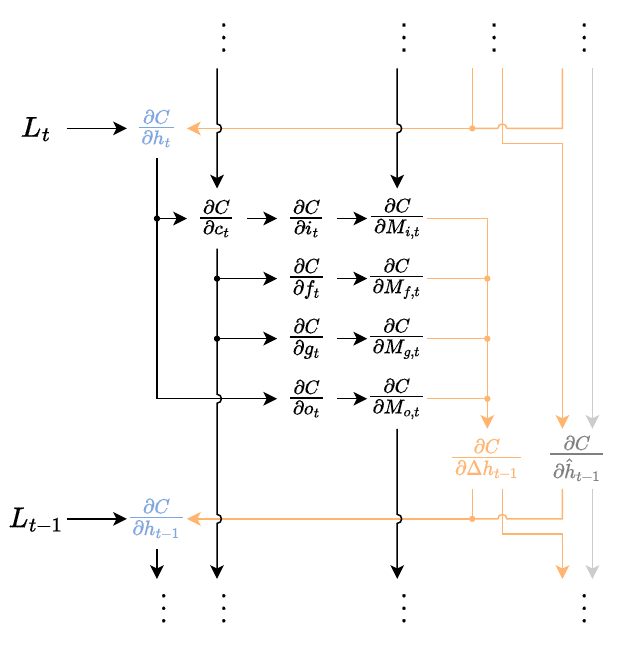}
        \caption{BP computation graph regarding \(h_t\).}
        \label{fig:cg_deltalstm_bp_h}
    \end{subfigure}
    \caption{Computation graphs of a Delta LSTMs layer.}
    \label{fig:cg_deltalstm}
\end{figure}

Figs.~\ref{fig:cg_deltalstm_fp_h} and \ref{fig:cg_deltalstm_bp_h} show the FP and BP computation graphs pertaining to the hidden state \(h_t\) of a Delta LSTM layer. The delta updating rules described by Eqs.~\eqref{eq:h_hat_it} and \eqref{eq:h_del_it} in Section~\ref{subsec:bptt_deltarnn} are also applied to the Delta LSTM. The pre-activation memory vector \(M_t\) consists of four components that correspond to the four gates of LSTM: input gate \(i_t\), forget gate \(f_t\), input modulation gate \(g_t\), and output gate \(o_t\). It also has a memory cell state \(c_t\). 

The FP equations of the Delta LSTM are shown below.

\begin{equation} \label{eq:lstm_M_it}
    M_{i,t} = \bm{W_{xi}} {\Delta x}_{t} + \bm{W_{hi}} {\Delta h}_{t-1} + M_{i,t-1}
\end{equation}
\begin{equation} \label{eq:lstm_M_ft}
    M_{f,t} = \bm{W_{xf}} {\Delta x}_{t} + \bm{W_{hf}} {\Delta h}_{t-1} + M_{f,t-1}
\end{equation}
\begin{equation} \label{eq:lstm_M_gt}
    M_{g,t} = \bm{W_{xg}} {\Delta x}_{t} + \bm{W_{hg}} {\Delta h}_{t-1} + M_{g,t-1}
\end{equation}
\begin{equation} \label{eq:lstm_M_ot}
    M_{o,t} = \bm{W_{xo}} {\Delta x}_{t} + \bm{W_{ho}} {\Delta h}_{t-1} + M_{o,t-1}
\end{equation}
\begin{equation}
    i_t = \sigma(M_{i,t})
\end{equation}
\begin{equation}
    f_t = \sigma(M_{f,t})
\end{equation}
\begin{equation}
    g_t = \tanh(M_{g,t})
\end{equation}
\begin{equation}
    o_t = \sigma(M_{o,t})
\end{equation}
\begin{equation}
    c_t = f_t \odot c_{t-1} + i_t \odot g_t
\end{equation}
\begin{equation}
    h_t = o_t \odot \tanh(c_t)
\end{equation}

Eqs.~\eqref{eq:lstm_M_it}-\eqref{eq:lstm_M_ot} can also be expressed by Eqs.~\eqref{eq:M_xh_t}-\eqref{eq:M_h_t} if we concatenate the weight matrices and the pre-activation memory vectors of the four gates. The memory vectors are initialized to the corresponding biases: \(M_{i,0} = b_i\), \(M_{f,0} = b_f\), \(M_{g,0} = b_g\), \(M_{o,0} = b_o\).

The BP equations are derived according to the computation graph (Fig.~\ref{fig:cg_deltalstm_bp_h}).

\begin{equation}
\begin{split}
    \pdv{C}{c_t} & = \pdv{C}{h_t} \pdv{h_t}{c_t} + \pdv{C}{c_{t+1}} \overline{\pdv{c_{t+1}}{c_t}} \\
                 & = \pdv{C}{h_t} \odot o_t \odot \tanh'(c_t) + \pdv{C}{c_{t+1}}
\end{split}
\end{equation}
\begin{equation}
\begin{split}
    \pdv{C}{M_{i,t}} & = \pdv{C}{c_t} \pdv{c_t}{i_t} \pdv{i_t}{M_{i,t}} + \pdv{C}{M_{i,t+1}} \overline{\pdv{M_{i,t+1}}{M_{i,t}}} \\
                     & = \pdv{C}{c_t} \odot g_t \odot \sigma'(M_{i,t}) + \pdv{C}{M_{i,t+1}}
\end{split}
\end{equation}
\begin{equation}
\begin{split}
    \pdv{C}{M_{f,t}} & = \pdv{C}{c_t} \pdv{c_t}{f_t} \pdv{f_t}{M_{f,t}} + \pdv{C}{M_{f,t+1}} \overline{\pdv{M_{f,t+1}}{M_{f,t}}} \\
                     & = \pdv{C}{c_t} \odot c_{t-1} \sigma'(M_{f,t}) + \pdv{C}{M_{f,t+1}}
\end{split}
\end{equation}
\begin{equation}
\begin{split}
    \pdv{C}{M_{g,t}} & = \pdv{C}{c_t} \pdv{c_t}{g_t} \pdv{g_t}{M_{g,t}} + \pdv{C}{M_{g,t+1}} \overline{\pdv{M_{g,t+1}}{M_{g,t}}} \\
                     & = \pdv{C}{c_t} \odot i_t \odot \tanh'(M_{g,t}) + \pdv{C}{M_{g,t+1}}
\end{split}
\end{equation}
\begin{equation}
\begin{split}
    \pdv{C}{M_{o,t}} & = \pdv{C}{h_t} \pdv{h_t}{o_t} \pdv{o_t}{M_{o,t}} + \pdv{C}{M_{o,t+1}} \overline{\pdv{M_{o,t+1}}{M_{o,t}}} \\
                     & = \pdv{C}{h_t} \odot \tanh(c_t) \odot \sigma'(M_{o,t}) + \pdv{C}{M_{o,t+1}}
\end{split}
\end{equation}

The term with an overline denotes the part of a partial derivative that only takes the direct connection into account. For example, \(\overline{\pdv{c_{t+1}}{c_t}}\) is the PD that comes from the path \(c_{t+1} \rightarrow c_t\) directly, excluding any other paths such as \(c_{t+1} \rightarrow i_{t+1} \rightarrow M_{i,t+1} \rightarrow {\Delta h}_t \rightarrow h_t \rightarrow c_t\).

The PD of the cost w.r.t. \({\Delta h}_{t-1}\) is:
\begin{equation} \label{eq:lstm_dc_ddh_ifgo}
\begin{aligned}
    \pdv{C}{{\Delta h}_{t-1}} ={} & \pdv{C}{M_{i,t}} \pdv{M_{i,t}}{{\Delta h}_{t-1}}
                                    + \pdv{C}{M_{f,t}} \pdv{M_{f,t}}{{\Delta h}_{t-1}} \\
                                  & + \pdv{C}{M_{h,t}} \pdv{M_{g,t}}{{\Delta h}_{t-1}}
                                    + \pdv{C}{M_{o,t}} \pdv{M_{o,t}}{{\Delta h}_{t-1}}
\end{aligned}
\end{equation}
The gradient vectors for the pre-activation memories of the four gates can also be concatenated into a longer vector \(\pdv{C}{M_t}\). Then Eq.~\eqref{eq:lstm_dc_ddh_ifgo} becomes:
\begin{equation} \label{eq:lstm_dc_ddh}
    \pdv{C}{{\Delta h}_{t-1}} = \pdv{C}{M_{t}} \pdv{M_{t}}{{\Delta h}_{t-1}} = \bm{W_{h}^{\intercal}} \pdv{C}{M_{t}}
\end{equation}

The PDs of the cost w.r.t. \({\hat h}_{t-1}\) and \(h_t\) are:
\begin{equation} \label{eq:lstm_dc_dhh}
\begin{split}
    \pdv{C}{{\hat h}_{t-1}} & = \pdv{C}{{\hat h}_{t}} \pdv{{\hat h}_{t}}{{\hat h}_{t-1}} + \pdv{C}{{\Delta h}_{t}} \pdv{{\Delta h}_{t}}{{\hat h}_{t-1}} \\
                            & = \pdv{C}{{\hat h}_{t}} \odot (1 - {m}_{h,t}) - \pdv{C}{{\Delta h}_{t}} \odot {m}_{h,t}
\end{split}
\end{equation}
\begin{equation} \label{eq:lstm_dc_dh}
\begin{split}
    \pdv{C}{h_{t}} & = \pdv{C}{{\hat h}_{t}} \pdv{{\hat h}_{t}}{h_{t}} + \pdv{C}{{\Delta h}_{t}} \pdv{{\Delta h}_{t}}{h_{t}} + \pdv{L_{t}}{h_{t}} \\
                   & = \pdv{C}{{\hat h}_{t}} \odot {m}_{h,t} + \pdv{C}{{\Delta h}_{t}} \odot {m}_{h,t} + \pdv{L_{t}}{h_{t}}
\end{split}
\end{equation}

The gradients of weight matrices obtained at the \(t\)-th timestep are given by:
\begin{align*}
    \bm{\frac{\partial C}{\partial W_{hi,t}}} = \pdv{C}{M_{i,t}} \pdv{M_{i,t}}{W_{hi,t}} \quad
    \bm{\frac{\partial C}{\partial W_{hf,t}}} = \pdv{C}{M_{f,t}} \pdv{M_{f,t}}{W_{hf,t}} \\
    \bm{\frac{\partial C}{\partial W_{hg,t}}} = \pdv{C}{M_{g,t}} \pdv{M_{g,t}}{W_{hg,t}} \quad
    \bm{\frac{\partial C}{\partial W_{ho,t}}} = \pdv{C}{M_{o,t}} \pdv{M_{o,t}}{W_{ho,t}}
\end{align*}
The gradients of biases are:
\begin{align*}
    \pdv{C}{b_i} = \pdv{C}{M_{i,0}} \qquad
    \pdv{C}{b_f} = \pdv{C}{M_{f,0}} \\
    \pdv{C}{b_g} = \pdv{C}{M_{g,0}} \qquad
    \pdv{C}{b_o} = \pdv{C}{M_{o,0}}
\end{align*}
By concatenating the four weight gradient matrices into \(\bm{\frac{\partial C}{\partial W_h}}\), we have:
\begin{equation} \label{eq:lstm_dc_dWh}
     \bm{\frac{\partial C}{\partial W_h}} = \sum_{t=1}^{T} {\pdv{C}{M_{t}} \pdv{M_{t}}{W_{h}}} = \sum_{t=1}^{T} {\pdv{C}{M_{t}} {\Delta h}_{t-1}^{\intercal}}
\end{equation}

It can be seen that Eqs.~\eqref{eq:lstm_dc_ddh}-\eqref{eq:lstm_dc_dWh} are the same as Eqs.~\eqref{eq:dc_ddh}-\eqref{eq:dc_dWh}.
Only the values at the indices of activated neurons in \(\pdv{C}{{\Delta h}_{t-1}}\) are used in Eqs.~\eqref{eq:lstm_dc_dhh} and \eqref{eq:lstm_dc_dh}, so Eq.~\eqref{eq:lstm_dc_ddh} can be made sparse using the mask vector \(m_{h,t}\). Eq.~\eqref{eq:lstm_dc_dWh} is also sparse since the term \(\pdv{C}{M_{t}}\) is multiplied with the sparse vector \({\Delta h}_{t-1}\). The sparsity in the gradient computations for the input \(x\) can be obtained in the same way.
Therefore, the temporal sparsity created in Delta LSTMs during FP also exist in the matrix multiplications during BP as shown in Eqs.~\eqref{eq:lstm_dc_ddh} and \eqref{eq:lstm_dc_dWh}. Other claims about Delta RNNs in Section~\ref{subsec:bptt_deltarnn} are also true for Delta LSTMs.

\subsection{Delta GRU BPTT Formulation}

\begin{figure}[t!]
    \centering
    \begin{subfigure}[t]{.45\textwidth}
        \centering
        \includegraphics[width=\linewidth]{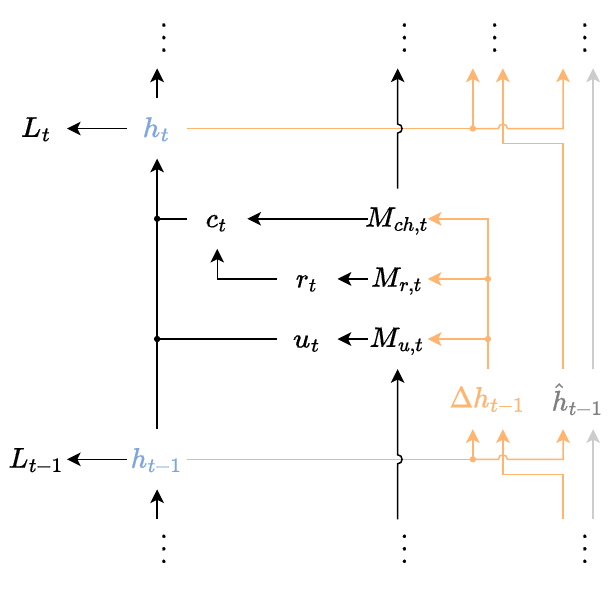}
        \caption{FP computation graph regarding \(h_t\).}
        \label{fig:cg_deltagru_fp_h}
    \end{subfigure}
    \hfill
    \begin{subfigure}[t]{.45\textwidth}
        \centering
        \includegraphics[width=\linewidth]{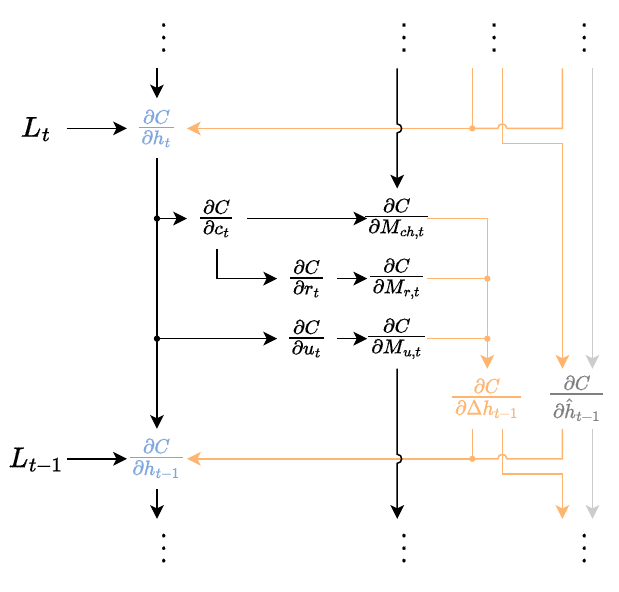}
        \caption{BP computation graph regarding \(h_t\).}
        \label{fig:cg_deltagru_bp_h}
    \end{subfigure}
    \caption{Computation graphs of a Delta GRU layer.}
    \label{fig:cg_deltagru}
\end{figure}

The GRU has a reset gate \(r\), an update gate \(u\), and a candidate activation vector \(c\). In Delta GRU we apply the delta updating rules described by Eqs.~\eqref{eq:h_hat_it} and \eqref{eq:h_del_it} in Section~\ref{subsec:bptt_deltarnn} to the input \(x\) and the hidden state \(h\). Figs.~\ref{fig:cg_deltagru_fp_h} and \ref{fig:cg_deltagru_bp_h} show the FP and BP computation graphs relevant to the hidden state \(h_t\) of a Delta GRU layer.

The FP equations of the Delta GRU are shown below.

\begin{equation} \label{eq:gru_M_rt}
    M_{r,t} = \bm{W_{xr}} {\Delta x}_{t} + \bm{W_{hr}} {\Delta h}_{t-1} + M_{r,t-1}
\end{equation}
\begin{equation} \label{eq:gru_M_ut}
    M_{u,t} = \bm{W_{xu}} {\Delta x}_{t} + \bm{W_{hu}} {\Delta h}_{t-1} + M_{u,t-1}
\end{equation}
\begin{equation} \label{eq:gru_M_cxt}
    M_{cx,t} = \bm{W_{xc}} {\Delta x}_{t} + M_{cx,t-1}
\end{equation}
\begin{equation} \label{eq:gru_M_cht}
    M_{ch,t} = \bm{W_{hc}} {\Delta h}_{t-1} + M_{ch,t-1}
\end{equation}
\begin{equation} \label{eq:gru_M_ct}
    M_{c,t} = M_{cx,t} + r_t \odot M_{ch,t}
\end{equation}
\begin{equation}
    r_t = \sigma(M_{r,t})
\end{equation}
\begin{equation}
    u_t = \sigma(M_{u,t})
\end{equation}
\begin{equation}
    c_t = \tanh(M_{c,t})
\end{equation}
\begin{equation} \label{eq:gru_h_t}
    h_t = (1 - u_t) \odot h_{t-1} + u_t \odot c_t
\end{equation}

The pre-activation memory vectors are initialized to their corresponding biases: \(M_{r,0} = b_r\), \(M_{u,0} = b_u\), \(M_{cx,0} = b_c\), \(M_{ch,0} = 0\).

The BP equations are derived according to the computation graph (Fig.~\ref{fig:cg_deltagru_bp_h}).

\begin{equation}
    \pdv{C}{c_t} = \pdv{C}{h_t} \pdv{h_t}{c_t}
                 = \pdv{C}{h_t} \odot u_t
\end{equation}
\begin{equation}
    \pdv{C}{M_{c,t}} = \pdv{C}{c_t} \pdv{c_t}{M_{c,t}}
                     = \pdv{C}{c_t} \odot \tanh'(M_{c,t})
\end{equation}
\begin{equation}
\begin{split}
    \pdv{C}{M_{ch,t}} & = \pdv{C}{M_{c,t}} \pdv{M_{c,t}}{M_{ch,t}} + \pdv{C}{M_{ch,t+1}} \overline{\pdv{M_{ch,t+1}}{M_{ch,t}}} \\
                      & = \pdv{C}{M_{c,t}} \odot r_t + \pdv{C}{M_{ch,t+1}}
\end{split}
\end{equation}
\begin{equation}
\begin{split}
    \pdv{C}{M_{r,t}} & = \pdv{C}{M_{c,t}} \pdv{M_{c,t}}{r_t} + \pdv{C}{M_{r,t+1}} \overline{\pdv{M_{r,t+1}}{M_{r,t}}} \\
                     & = \pdv{C}{M_{c,t}} \odot M_{ch,t} + \pdv{C}{M_{r,t+1}}
\end{split}
\end{equation}
\begin{equation}
\begin{split}
    \pdv{C}{M_{u,t}} & = \pdv{C}{h_t} \pdv{h_t}{u_t} \pdv{u_t}{M_{u,t}} + \pdv{C}{M_{u,t+1}} \overline{\pdv{M_{u,t+1}}{M_{u,t}}} \\
                 ={} & \pdv{C}{h_t} \odot (c_t - h_{t-1}) \odot \sigma'(M_{u,t}) + \pdv{C}{M_{u,t+1}}
\end{split}
\end{equation}

The PD of the cost w.r.t. \({\Delta h}_{t-1}\) is:
\begin{equation} \label{eq:gru_dc_ddh_ruc}
\begin{aligned}
    \pdv{C}{{\Delta h}_{t-1}} = {} & \pdv{C}{M_{u,t}} \pdv{M_{u,t}}{{\Delta h}_{t-1}}
                                     + \pdv{C}{M_{r,t}} \pdv{M_{r,t}}{{\Delta h}_{t-1}} \\
                                   & + \pdv{C}{M_{ch,t}} \pdv{M_{ch,t}}{{\Delta h}_{t-1}}
\end{aligned}
\end{equation}
The gradient vectors \(\pdv{C}{M_{r,t}}\), \(\pdv{C}{M_{u,t}}\), and \(\pdv{C}{M_{ch,t}}\) can be concatenated into a longer vector \(\pdv{C}{M_{h,t}}\). Their weight matrices \(\bm{W_{hr}}\), \(\bm{W_{hu}}\), and \(\bm{W_{hc}}\) can also be concatenated as \(\bm{W_{h}^{\intercal}}\). Then Eq.~\eqref{eq:gru_dc_ddh_ruc} becomes:
\begin{equation} \label{eq:gru_dc_ddh}
    \pdv{C}{{\Delta h}_{t-1}} = \pdv{C}{M_{h,t}} \pdv{M_{h,t}}{{\Delta h}_{t-1}} = \bm{W_{h}^{\intercal}} \pdv{C}{M_{h,t}}
\end{equation}

The PDs of the cost w.r.t. \({\hat h}_{t-1}\) and \(h_t\) are:
\begin{equation} \label{eq:gru_dc_dhh}
\begin{split}
    \pdv{C}{{\hat h}_{t-1}} ={} & \pdv{C}{{\hat h}_{t}} \pdv{{\hat h}_{t}}{{\hat h}_{t-1}} + \pdv{C}{{\Delta h}_{t}} \pdv{{\Delta h}_{t}}{{\hat h}_{t-1}} \\
                            ={} & \pdv{C}{{\hat h}_{t}} \odot (1 - {m}_{h,t}) - \pdv{C}{{\Delta h}_{t}} \odot {m}_{h,t}
\end{split}
\end{equation}
\begin{equation} \label{eq:gru_dc_dh}
\begin{split}
    \pdv{C}{h_{t}} ={} & \pdv{C}{{\hat h}_{t}} \pdv{{\hat h}_{t}}{h_{t}} + \pdv{C}{{\Delta h}_{t}} \pdv{{\Delta h}_{t}}{h_{t}} \\
                       & + \pdv{C}{h_{t+1}} \overline{\pdv{h_{t+1}}{h_{t}}} + \pdv{L_{t}}{h_{t}} \\
                   ={} & \pdv{C}{{\hat h}_{t}} \odot {m}_{h,t} + \pdv{C}{{\Delta h}_{t}} \odot {m}_{h,t} \\
                       & + \pdv{C}{h_{t+1}} \odot (1 - u_{t+1}) + \pdv{L_{t}}{h_{t}}
\end{split}
\end{equation}

The gradients of weight matrices obtained at the \(t\)-th timestep are given by:
\begin{gather*}
    \bm{\frac{\partial C}{\partial W_{hr,t}}} = \pdv{C}{M_{r,t}} \pdv{M_{r,t}}{W_{hr,t}} \quad
    \bm{\frac{\partial C}{\partial W_{hu,t}}} = \pdv{C}{M_{u,t}} \pdv{M_{u,t}}{W_{hu,t}} \\
    \bm{\frac{\partial C}{\partial W_{hc,t}}} = \pdv{C}{M_{ch,t}} \pdv{M_{ch,t}}{W_{hc,t}}
\end{gather*}
The gradients of biases are:
\begin{equation*}
    \pdv{C}{b_r} = \pdv{C}{M_{r,0}} \quad
    \pdv{C}{b_u} = \pdv{C}{M_{u,0}} \quad
    \pdv{C}{b_c} = \pdv{C}{M_{cx,0}}
\end{equation*}
By concatenating the weight gradient matrices \(\bm{\frac{\partial C}{\partial W_{hr}}}\), \(\bm{\frac{\partial C}{\partial W_{hu}}}\), and \(\bm{\frac{\partial C}{\partial W_{hc}}}\) into \(\bm{\frac{\partial C}{\partial W_h}}\), we have:
\begin{equation} \label{eq:gru_dc_dWh}
     \bm{\frac{\partial C}{\partial W_h}} = \sum_{t=1}^{T} {\pdv{C}{M_{h,t}} \pdv{M_{h,t}}{W_{h}}} = \sum_{t=1}^{T} {\pdv{C}{M_{h,t}} {\Delta h}_{t-1}^{\intercal}}
\end{equation}

It can be seen that Eqs.~\eqref{eq:gru_dc_ddh}-\eqref{eq:gru_dc_dWh} are essentially the same as Eqs.~\eqref{eq:dc_ddh}-\eqref{eq:dc_dWh}, except that there is an additional term in Eq.~\eqref{eq:gru_dc_dh} derived from the self-recursion of \(h_t\) in Eq.~\eqref{eq:gru_h_t}.
But the values at the indices of inactivated neurons in \(\pdv{C}{{\Delta h}_{t-1}}\) are still unused in BP, so Eq.~\eqref{eq:gru_dc_ddh} can still be sparsified with the mask vector \(m_{h,t}\). Eq.~\eqref{eq:gru_dc_dWh} is also sparse as the term \(\pdv{C}{M_{h,t}}\) is multiplied with the sparse vector \({\Delta h}_{t-1}\). The sparsity in the gradient computations for the input \(x\) can be obtained in the same manner.
Therefore, the temporal sparsity created in Delta GRUs during FP also exist in the matrix multiplications during BP as shown in Eqs.~\eqref{eq:gru_dc_ddh} and \eqref{eq:gru_dc_dWh}. Other claims about Delta RNNs in Section~\ref{subsec:bptt_deltarnn} are also true for Delta GRUs.



\end{appendices}

%% file: arxiv.bbl
\begin{thebibliography}{24}
\providecommand{\natexlab}[1]{#1}

\bibitem[{Aimar et~al.(2019)Aimar, Mostafa, Calabrese, Rios-Navarro,
  Tapiador-Morales, Lungu, Milde, Corradi, Linares-Barranco, Liu, and
  Delbruck}]{aimar2019}
Aimar, A.; Mostafa, H.; Calabrese, E.; Rios-Navarro, A.; Tapiador-Morales, R.;
  Lungu, I.-A.; Milde, M.~B.; Corradi, F.; Linares-Barranco, A.; Liu, S.-C.;
  and Delbruck, T. 2019.
\newblock {NullHop}: A Flexible Convolutional Neural Network Accelerator Based
  on Sparse Representations of Feature Maps.
\newblock \emph{IEEE Transactions on Neural Networks and Learning Systems},
  30(3): 644--656.

\bibitem[{Campos et~al.(2017)Campos, Jou, Gir{\'o}-i Nieto, Torres, and
  Chang}]{campos2017skip}
Campos, V.; Jou, B.; Gir{\'o}-i Nieto, X.; Torres, J.; and Chang, S.-F. 2017.
\newblock Skip {RNN}: Learning to skip state updates in recurrent neural
  networks.
\newblock \emph{arXiv preprint arXiv:1708.06834}.

\bibitem[{Chen, Blair, and
  Cong(2022)}]{Chen2022-cong-lstm-sparse-bit-shift-macs}
Chen, Z.; Blair, H.~T.; and Cong, J. 2022.
\newblock {Energy-Efficient} {LSTM} Inference Accelerator for {Real-Time}
  Causal Prediction.
\newblock \emph{ACM Trans. Des. Automat. Electron. Syst.}, 27(5): 1--19.

\bibitem[{Cho et~al.(2014)Cho, van Merrienboer, Gulcehre, Bougares, Schwenk,
  and Bengio}]{Cho2014}
Cho, K.; van Merrienboer, B.; Gulcehre, C.; Bougares, F.; Schwenk, H.; and
  Bengio, Y. 2014.
\newblock Learning phrase representations using RNN encoder-decoder for
  statistical machine translation.
\newblock In \emph{Conference on Empirical Methods in Natural Language
  Processing (EMNLP 2014)}.

\bibitem[{Gao, Delbruck, and Liu(2022)}]{gao2022spartus}
Gao, C.; Delbruck, T.; and Liu, S.-C. 2022.
\newblock {Spartus: A 9.4 TOp/s FPGA-based LSTM} accelerator exploiting
  spatio-temporal sparsity.
\newblock \emph{IEEE Transactions on Neural Networks and Learning Systems}.

\bibitem[{Gao et~al.(2018)Gao, Neil, Ceolini, Liu, and
  Delbruck}]{GaoDeltaRNN2018}
Gao, C.; Neil, D.; Ceolini, E.; Liu, S.-C.; and Delbruck, T. 2018.
\newblock {DeltaRNN: A} Power-efficient Recurrent Neural Network Accelerator.
\newblock In \emph{Proceedings of the 2018 ACM/SIGDA International Symposium on
  Field-Programmable Gate Arrays}, FPGA '18, 21--30. New York, NY, USA: ACM.
\newblock ISBN 978-1-4503-5614-5.

\bibitem[{Gao et~al.(2020)Gao, Rios-Navarro, Chen, Liu, and
  Delbruck}]{edgedrnn}
Gao, C.; Rios-Navarro, A.; Chen, X.; Liu, S.-C.; and Delbruck, T. 2020.
\newblock {EdgeDRNN}: Recurrent Neural Network Accelerator for Edge Inference.
\newblock \emph{IEEE Journal on Emerging and Selected Topics in Circuits and
  Systems}, 10(4): 419--432.

\bibitem[{Giraldo, Jain, and Verhelst(2021)}]{giraldo2021efficient}
Giraldo, J. S.~P.; Jain, V.; and Verhelst, M. 2021.
\newblock Efficient execution of temporal convolutional networks for embedded
  keyword spotting.
\newblock \emph{IEEE Transactions on Very Large Scale Integration (VLSI)
  Systems}, 29(12): 2220--2228.

\bibitem[{Hochreiter and Schmidhuber(1997)}]{lstm_hoch97}
Hochreiter, S.; and Schmidhuber, J. 1997.
\newblock Long Short-Term Memory.
\newblock \emph{Neural Comput.}, 9(8): 1735--1780.

\bibitem[{Horowitz(2014)}]{horowitz20141}
Horowitz, M. 2014.
\newblock 1.1 Computing's energy problem (and what we can do about it).
\newblock In \emph{2014 IEEE International Solid-State Circuits Conference
  Digest of Technical Papers (ISSCC)}, 10--14.

\bibitem[{Hunter, Spracklen, and Ahmad(2022)}]{hunter2022two}
Hunter, K.; Spracklen, L.; and Ahmad, S. 2022.
\newblock Two sparsities are better than one: unlocking the performance
  benefits of sparse--sparse networks.
\newblock \emph{Neuromorphic Computing and Engineering}, 2(3): 034004.

\bibitem[{Kadetotad et~al.(2020)Kadetotad, Yin, Berisha, and
  {others}}]{Kadetotad2020-seo-lstm-cg-sparsity-jssc}
Kadetotad, D.; Yin, S.; Berisha, V.; and {others}. 2020.
\newblock An 8.93 {TOPS/W} {LSTM} recurrent neural network accelerator
  featuring hierarchical coarse-grain sparsity for on-device speech
  recognition.
\newblock \emph{IEEE J. Solid-State Circuits}.

\bibitem[{Kingma and Ba(2015)}]{KingBa15}
Kingma, D.; and Ba, J. 2015.
\newblock Adam: A Method for Stochastic Optimization.
\newblock In \emph{International Conference on Learning Representations
  (ICLR)}. San Diega, CA, USA.

\bibitem[{Krueger et~al.(2017)Krueger, Maharaj, Kramár, Pezeshki, Ballas, Ke,
  Goyal, Bengio, Courville, and Pal}]{krueger2017zoneout}
Krueger, D.; Maharaj, T.; Kramár, J.; Pezeshki, M.; Ballas, N.; Ke, N.~R.;
  Goyal, A.; Bengio, Y.; Courville, A.; and Pal, C. 2017.
\newblock Zoneout: Regularizing {RNNs} by Randomly Preserving Hidden
  Activations.
\newblock arXiv:1606.01305.

\bibitem[{Lindmar, Gao, and
  Liu(2022)}]{Lindmar2022-liu-targeted-dropout-lstm-training}
Lindmar, J.~H.; Gao, C.; and Liu, S.-C. 2022.
\newblock Intrinsic sparse {LSTM} using structured targeted dropout for
  efficient hardware inference.
\newblock In \emph{2022 {IEEE} 4th International Conference on Artificial
  Intelligence Circuits and Systems ({AICAS})}, 126--129.

\bibitem[{Lugosch et~al.(2019)Lugosch, Ravanelli, Ignoto, Tomar, and
  Bengio}]{Lugosch2019}
Lugosch, L.; Ravanelli, M.; Ignoto, P.; Tomar, V.~S.; and Bengio, Y. 2019.
\newblock {Speech Model Pre-Training for End-to-End Spoken Language
  Understanding}.
\newblock In \emph{Proc. Interspeech 2019}, 814--818.

\bibitem[{Neil et~al.(2017)Neil, Lee, Delbr{\"{u}}ck, and Liu}]{neil2016delta}
Neil, D.; Lee, J.; Delbr{\"{u}}ck, T.; and Liu, S. 2017.
\newblock Delta Networks for Optimized Recurrent Network Computation.
\newblock In \emph{Proceedings of the 34th International Conference on Machine
  Learning, {ICML} 2017, Sydney, NSW, Australia, 6-11 August 2017}, 2584--2593.

\bibitem[{O'Connor and Welling(2016)}]{o2016sigma}
O'Connor, P.; and Welling, M. 2016.
\newblock Sigma delta quantized networks.
\newblock \emph{arXiv preprint arXiv:1611.02024}.

\bibitem[{Perez-Nieves and Goodman(2021)}]{perez-nieves2021sparse}
Perez-Nieves, N.; and Goodman, D. F.~M. 2021.
\newblock Sparse Spiking Gradient Descent.
\newblock In Beygelzimer, A.; Dauphin, Y.; Liang, P.; and Vaughan, J.~W., eds.,
  \emph{Advances in Neural Information Processing Systems}.

\bibitem[{Rebuffi et~al.(2017)Rebuffi, Kolesnikov, Sperl, and
  Lampert}]{rebuffi2017icarl}
Rebuffi, S.-A.; Kolesnikov, A.; Sperl, G.; and Lampert, C.~H. 2017.
\newblock {iCARL: I}ncremental classifier and representation learning.
\newblock In \emph{Proceedings of the IEEE Conference on Computer Vision and
  Pattern Recognition}, 2001--2010.

\bibitem[{Shan et~al.(2020)Shan, Yang, Wang, Lu, Cai, Zhu, Xu, Wu, Shi, and
  Yang}]{shan2020510}
Shan, W.; Yang, M.; Wang, T.; Lu, Y.; Cai, H.; Zhu, L.; Xu, J.; Wu, C.; Shi,
  L.; and Yang, J. 2020.
\newblock A 510-nW wake-up keyword-spotting chip using serial-FFT-based MFCC
  and binarized depthwise separable CNN in 28-nm CMOS.
\newblock \emph{IEEE Journal of Solid-State Circuits}, 56(1): 151--164.

\bibitem[{Srivastava et~al.(2019)Srivastava, Kadetotad, Yin, Berisha,
  Chakrabarti, and Seo}]{Srivastava2019-Seo-structured-RNN-quantization}
Srivastava, G.; Kadetotad, D.; Yin, S.; Berisha, V.; Chakrabarti, C.; and Seo,
  J.-S. 2019.
\newblock Joint optimization of quantization and structured sparsity for
  compressed deep neural networks.
\newblock In \emph{{ICASSP} 2019 - 2019 {IEEE} International Conference on
  Acoustics, Speech and Signal Processing ({ICASSP})}, 1393--1397. IEEE.
\newblock ISBN 9781479981311.

\bibitem[{Subramoney et~al.(2022)Subramoney, Nazeer, Sch{\"o}ne, Mayr, and
  Kappel}]{subramoneyefficient}
Subramoney, A.; Nazeer, K.~K.; Sch{\"o}ne, M.; Mayr, C.; and Kappel, D. 2022.
\newblock Efficient recurrent architectures through activity sparsity and
  sparse back-propagation through time.
\newblock In \emph{The Eleventh International Conference on Learning
  Representations}.

\bibitem[{Warden(2018)}]{warden2018speech}
Warden, P. 2018.
\newblock Speech commands: A dataset for limited-vocabulary speech recognition.
\newblock \emph{arXiv preprint arXiv:1804.03209}.

\end{thebibliography}
